\documentclass[12pt,a4paper]{article}

\usepackage{natbib}
\bibpunct{(}{)}{;}{a}{,}{,}

\usepackage{graphicx}
\usepackage{subfigure}
\graphicspath{{figure/},{illustration/}}

\usepackage{algorithm,algorithmic}

\usepackage{amsmath,amsthm,amssymb,mathrsfs}

\newtheorem{theorem}{Theorem}

\theoremstyle{remark}

\DeclareMathOperator{\tr}{\mathrm{tr}}
\DeclareMathOperator{\vect}{\mathrm{vec}}
\DeclareMathOperator{\sign}{\mathrm{sign}}

\DeclareMathOperator*{\argmax}{\mathrm{arg\,max}}

\newcommand{\T}{{\hspace{-0.25ex}\top\hspace{-0.25ex}}}
\newcommand{\ST}{\mathrm{s.t.}}
\newcommand{\bR}{\mathbb{R}}
\newcommand{\bE}{\mathbb{E}}

\newcommand{\zero}{\boldsymbol{0}}
\newcommand{\one}{\boldsymbol{1}}
\newcommand{\ve}{\boldsymbol{e}}
\newcommand{\vh}{\boldsymbol{h}}

\newcommand{\vr}{\boldsymbol{r}}

\newcommand{\vv}{\boldsymbol{v}}

\newcommand{\vy}{\boldsymbol{y}}
\newcommand{\vz}{\boldsymbol{z}}

\newcommand{\cC}{\mathcal{C}}

\newcommand{\cE}{\mathcal{E}}

\newcommand{\cH}{\mathcal{H}}

\newcommand{\cN}{\mathcal{N}}

\newcommand{\cP}{\mathcal{P}}

\newcommand{\cU}{\mathcal{U}}
\newcommand{\cV}{\mathcal{V}}

\newcommand{\cX}{\mathcal{X}}
\newcommand{\cY}{\mathcal{Y}}

\begin{document}

\title{\vspace*{-33mm}%
    \begin{flushleft}
    \normalsize\sl Submitted to ICML 2014.
    \end{flushleft}
    \vspace*{2em}%
    Transductive Learning with Multi-class\\Volume Approximation%
}

\author{Gang Niu\\
    Tokyo Institute of Technology, Japan\\
    gang@sg.cs.titech.ac.jp\\[1ex]
Bo Dai\\
    Georgia Institute of Technology, USA\\
    bohr.dai@gmail.com\\[1ex]
Marthinus Christoffel du Plessis\\
    Tokyo Institute of Technology, Japan\\
    christo@sg.cs.titech.ac.jp\\[1ex]
Masashi Sugiyama\\
    Tokyo Institute of Technology, Japan\\
    sugi@cs.titech.ac.jp
}

\date{}

\maketitle
\thispagestyle{myheadings}
\markright{}

\begin{abstract}\noindent%
  Given a hypothesis space, the \emph{large volume principle} by Vladimir Vapnik prioritizes equivalence classes according to their volume in the hypothesis space. The \emph{volume approximation} has hitherto been successfully applied to binary learning problems. In this paper, we extend it naturally to a more general definition which can be applied to several transductive problem settings, such as \emph{multi-class}, \emph{multi-label} and \emph{serendipitous} learning. Even though the resultant learning method involves a non-convex optimization problem, the globally optimal solution is almost surely unique and can be obtained in $O(n^3)$ time. We theoretically provide stability and error analyses for the proposed method, and then experimentally show that it is promising.
\end{abstract}

\pagestyle{myheadings}
\markright{123}

\section{Introduction}

The history of the \emph{large volume principle} (LVP) goes back to the early age of the statistical learning theory when \citet{vapnik82} introduced it for the case of hyperplanes. But it did not gain much attention until a creative approximation was proposed in \citet{elyaniv08} to implement LVP for the case of soft response vectors. From then on, it has been applied to various binary learning problems successfully, such as binary transductive learning \citep{elyaniv08}, binary clustering \citep{niu13}, and outlier detection \citep{li13}.

\begin{figure}[t]
  \centering
  \includegraphics[width=0.6\columnwidth]{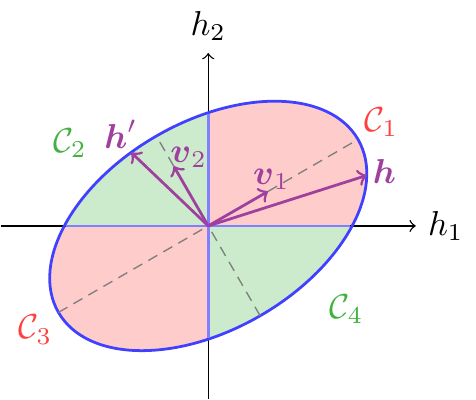}
  \caption{The large volume principle and its approximation.}
  \label{fig:lvp}
\end{figure}

LVP is a learning-theoretic principle which views learning as \emph{hypothesis selecting} from a certain \emph{hypothesis space} $\cH$. Despite the form of the hypothesis, $\cH$ can always be partitioned into a finite number of equivalence classes after we observe certain data, where an \emph{equivalence class} is a set of hypotheses that generate the same labeling of the observed data. LVP, as one of the learning-theoretic principles from the statistical learning theory, prioritizes those equivalence classes according to the volume they occupy in $\cH$. See the illustration in Figure~\ref{fig:lvp}: The blue ellipse represents $\cH$, and it is partitioned into $\cC_1,\ldots,\cC_4$ each occupying a quadrant of the Cartesian coordinate system $\bR^2$ intersected with $\cH$; LVP claims that $\cC_1$ and $\cC_3$ are more preferable than $\cC_2$ and $\cC_4$, since $\cC_1$ and $\cC_3$ have larger volume than $\cC_2$ and $\cC_4$.

In practice, the hypothesis space $\cH$ cannot be as simple as in Figure~\ref{fig:lvp}. It frequently locates in very high-dimensional spaces where exact or even quantifiable \emph{volume estimation} is challenging. Therefore, \citet{elyaniv08} proposed a \emph{volume approximation} to bypass the volume estimation. Instead of focusing on the equivalence classes of $\cH$, it directly focuses on the hypotheses in $\cH$ since learning is regarded as hypothesis selecting in LVP. It defines $\cH$ via an ellipsoid, measures the angles from hypotheses to the principal axes of $\cH$, and then prefers hypotheses near the long principal axes to those near the short ones. This manner is reasonable, since the long principal axes of $\cH$ lie in large-volume regions. In Figure~\ref{fig:lvp}, $\vh$ and $\vh'$ are two hypotheses and $\vv_1$/$\vv_2$ is the long/short principal axis; LVP advocates that $\vh$ is more preferable than $\vh'$ as $\vh$ is close to $\vv_1$ and $\vh'$ is close to $\vv_2$. We can adopt this volume approximation to regularize our loss function, which has been demonstrated helpful for various binary learning problems.

Nevertheless, the volume approximation in \citet{elyaniv08} only fits binary learning problem settings in spite of its potential advantages. In this paper, we naturally extend it to a more general definition that can be applied to some transductive problem settings including but not limited to \emph{multi-class} learning \citep{zhou03}, \emph{multi-label} learning \citep{kong13}, and \emph{serendipitous} learning \citep{zhang11}. We adopt the same strategy as \citet{elyaniv08}: For $n$ data and $c$ labels, a hypothesis space is defined in $\bR^{n\times c}$ and linked to an ellipsoid in $\bR^{nc}$, such that the equivalence classes and the volume approximation can be defined accordingly. Similarly to the binary volume approximation, our approach is also distribution free, that is, the labeled and unlabeled data do not necessarily share the same marginal distribution. This advantage of transductive learning over (semi-supervised) inductive learning is especially useful for serendipitous problems where the labeled and unlabeled data must not be identically distributed.

We name the learning method which realizes the proposed multi-class volume approximation \emph{multi-class approximate volume regularization} (MAVR). It involves a non-convex optimization problem, but the globally optimal solution is almost surely unique and accessible in $O(n^3)$ time following \citet{forsythe65}. Moreover, we theoretically provide stability and error analyses for MAVR, as well as experimentally compare it to two state-of-the-art methods in \citet{zhou03} and \citet{belkin06} using USPS, MNIST, 20Newsgroups and Isolet.

The rest of this paper is organized as follows. In Section~2 the binary volume approximation is reviewed, and in Section~3 the multi-class volume approximation is derived. In Section~4, we develop and analyze MAVR. Finally, the experimental results are in Section~5.

\section{Binary Volume Approximation}

The binary volume approximation in \citet{elyaniv08} involves a few key concepts: The soft response vector, the hypothesis space and the equivalence class, and the power and volume of equivalence classes. We review the concepts in this section for later use in the next section.

Suppose that $\cX$ is the domain of input data, and most often but not necessarily, $\cX\subset\bR^d$ where $d$ is a natural number. Given a set of $n$ data $X_n=\{x_1,\ldots,x_n\}$ where $x_i\in\cX$, a \emph{soft response vector} is an $n$-dimensional vector
\begin{equation}
\label{eq:srv}%
\vh:=(h_1,\ldots,h_n)^\T\in\bR^n,
\end{equation}
so that $h_i$ stands for a soft or confidence-rated label of $x_i$. For binary transductive learning problems, a soft response vector $\vh$ suggests that $x_i$ is from the positive class if $h_i>0$, $x_i$ is from the negative class if $h_i<0$, and the above two cases are equally possible if $h_i=0$.

A \emph{hypothesis space} is a collection of hypotheses. The volume approximation requires a symmetric positive-definite matrix $Q\in\bR^{n\times n}$ which contains the pairwise information about $X_n$. Consider the hypothesis space
\begin{equation}
\label{eq:hypo-spa-bin}%
\cH_Q:=\{\vh\mid\vh^\T Q\vh\le1\},
\end{equation}
where the hypotheses are soft response vectors. The set of sign vectors $\{\sign(\vh)\mid\vh\in\cH_Q\}$ contains all of $N=2^n$ possible dichotomies of $X_n$, and $\cH_Q$ can be partitioned into a finite number of \emph{equivalence classes} $\cC_1,\ldots,\cC_N$, such that for fixed $k$, all hypotheses in $\cC_k$ will generate the same labeling of $X_n$.

Then, in statistical learning theory, the \emph{power} of an equivalence class $\cC_k$ is defined as the probability mass of all hypotheses in it \citep[p.~708]{vapnik98}, i.e.,
\[ \cP(\cC_k):=\int_{\cC_k}p(\vh)\mathrm{d}\vh,\quad k=1,\ldots,N, \]
where $p(\vh)$ is the underlying probability density of $\vh$ over $\cH_Q$. The hypotheses in $\cC_k$ which has a large power should be preferred according to \citet{vapnik98}.

When no specific domain knowledge is available (i.e., $p(\vh)$ is unknown), it would be natural to assume the continuous uniform distribution $p(\vh)=1/\sum_{k=1}^N\cV(\cC_k)$, where
\[ \cV(\cC_k):=\int_{\cC_k}\mathrm{d}\vh,\quad k=1,\ldots,N, \]
is the \emph{volume} of $\cC_k$. That is, the volume of an equivalence class is defined as the geometric volume of all hypotheses in it. As a result, $\cP(\cC_k)$ is proportional to $\cV(\cC_k)$, and the larger the value $\cV(\cC_k)$ is, the more confident we are of the hypotheses chosen from $\cC_k$.

However, it is very hard to accurately compute the geometric volume of even a single convex body in $\bR^n$, let alone all $2^n$ convex bodies, so \citet{elyaniv08} introduced an efficient approximation. Let $\lambda_1\le\cdots\le\lambda_n$ be the eigenvalues of $Q$, and $\vv_1,\ldots,\vv_n$ be the associated orthonormal eigenvectors. Actually, the hypothesis space $\cH_Q$ in Eq.~\eqref{eq:hypo-spa-bin} is geometrically an origin-centered ellipsoid in $\bR^n$ with $\vv_i$ and $1/\sqrt{\lambda_i}$ as the direction and length of its $i$-th principal axis. Note that a small angle from a hypothesis $\vh$ in $\cC_k$ to some $\vv_i$ with a small/large index $i$ (i.e., a long/short principal axis) implies that $\cV(\cC_k)$ is large/small (cf.~Figure~\ref{fig:lvp}). Based on this crucial observation, we define
\begin{equation}
\label{eq:volume-bin}%
V(\vh):=\sum_{i=1}^n\lambda_i\left(\frac{\vh^\T\vv_i}{\|\vh\|_2}\right)^2 =\frac{\vh^\T Q\vh}{\|\vh\|_2^2},
\end{equation}
where $\vh^\T\vv_i/\|\vh\|_2$ means the cosine of the angle between $\vh$ and $\vv_i$. We subsequently expect $V(\vh)$ to be small when $\vh$ lies in a large-volume equivalence class, and conversely to be large when $\vh$ lies in a small-volume equivalence class.

\section{Multi-class Volume Approximation}

In this section, we propose a more general multi-class volume approximation that fits for several problem settings.

\subsection{Problem settings}

Recall the setting of binary transductive problems \citep[p.~341]{vapnik98}. A fixed set $X_n=\{x_1,\ldots,x_n\}$ of $n$ points from $\cX$ is observed, and the labels $y_1,\ldots,y_n\in\{-1,+1\}$ of these points are also fixed but unknown. A subset $X_l\subset X_n$ of size $l$ is picked uniformly at random, and then $y_i$ is revealed if $x_i\in X_l$. We call $S_l=\{(x_i,y_i)\mid x_i\in X_l\}$ the labeled data and $X_u=X_n\setminus X_l$ the unlabeled data. Using $S_l$ and $X_u$, the goal is to predict $y_i$ of $x_i\in X_u$ (while any unobserved $x\in\cX\setminus X_n$ is currently left out of account).

A slight modification suffices to extend the setting. Instead of $y_1,\ldots,y_n\in\{-1,+1\}$, we assume that $y_1,\ldots,y_n\in\cY$ where $\cY=\{1,\ldots,c\}$ is the domain of labels and $c$ is a natural number. Though the binary setting is popular, this multi-class setting has been studied in just a few previous works such as \citet{szummer01} and \citet{zhou03}. Without loss of generality, we assume that each of the $c$ labels possesses some labeled data.

In addition, it would be a multi-label setting, if $y_1,\ldots,y_n$ $\subseteq\cY$ with $\cY=\{1,\ldots,c\}$ where each $y_i$ is a label set, or if $y_1,\ldots,y_n\in\cY$ with $\cY=\{-1,0,1\}^c$ where each $y_i$ is a label vector. To the best of our knowledge, the former setting has been studied only in \citet{kong13} and the latter setting has not been studied yet. The latter setting is more general, since the former one requires labeled data to be fully labeled, while the latter one allows labeled data to be partially labeled. A huge challenge of multi-label problems is that some label sets or label vectors might have no labeled data \citep{kong13}, since there are $2^c$ possible label sets and $3^c$ possible label vectors.

A more challenging serendipitous setting which is a multi-class setting but some labels have no labeled data has been studied in \citet{zhang11}. Let $\cY_l=\{y_i\mid x_i\in X_l\}$ and $\cY_u=\{y_i\mid x_i\in X_u,y_i\not\in\cY_l\}$, then we have $\#\cY_u\ge1$ where $\#$ measures the cardinality. It is still solvable when $\#\cY_u=1$ if a special label of outliers is allowed and when $\#\cY_u>1$ as a combination of classification and clustering problems. \citet{zhang11} is the unique previous work which successfully dealt with $\#\cY_u=2$ and $\#\cY_u=3$.

\subsection{Definitions}

The multi-class volume approximation to be proposed can handle all the problem settings discussed so far in a unified manner. In order to extend the binary definitions, we need only to extend the hypothesis and the hypothesis space.

To begin with, we allocate a soft response vector in Eq.~\eqref{eq:srv} for each of the $c$ labels:
\[ \vh_1=(h_{1,1},\ldots,h_{n,1})^\T,\ldots,\vh_c=(h_{1,c},\ldots,h_{n,c})^\T. \]
The value $h_{i,j}$ is a soft or confidence-rated label of $x_i$ concerning the $j$-th label and it suggests that
\begin{itemize}
  \setlength{\itemsep}{0em}%
  \item $x_i$ should possess the $j$-th label, if $h_{i,j}>0$;
  \item $x_i$ should not possess the $j$-th label, if $h_{i,j}<0$;
  \item the above two cases are equally possible, if $h_{i,j}=0$.
\end{itemize}
For multi-class and serendipitous problems, $y_i$ is predicted by $\hat{y}_i=\argmax_jh_{i,j}$. For multi-label problems, we need a threshold $T_h$ that is either preset or learned since usually positive and negative labels are imbalanced, and $y_i$ can be predicted by $\hat{y}_i=\{j\mid h_{i,j}\ge T_h\}$; or we can use the label set prediction methods proposed in \citet{kong13}.

Then, a \emph{soft response matrix} as our transductive hypothesis is an $n$-by-$c$ matrix defined by
\begin{equation}
\label{eq:srm}%
H=(\vh_1,\ldots,\vh_c)\in\bR^{n\times c},
\end{equation}
and a \emph{stacked soft response vector} as an equivalent hypothesis is an $nc$-dimensional vector defined by
\[ \vh=\vect(H)=(\vh_1^\T,\ldots,\vh_c^\T)^\T\in\bR^{nc}, \]
where $\vect(H)$ is the vectorization of $H$ formed by stacking its columns into a single vector.

As the binary definition of the hypothesis space, a symmetric positive-definite matrix $Q\in\bR^{n\times n}$ which contains the pairwise information about $X_n$ is provided, and we assume further that a symmetric positive-definite matrix $P\in\bR^{c\times c}$ which contains the pairwise information about $\cY$ is available. Consider the hypothesis space
\begin{equation}
\label{eq:hypo-spa-mul}%
\cH_{P,Q} :=\{H\mid\tr(H^\T QHP)\le1\},
\end{equation}
where the hypotheses are soft response matrices. Let $P\otimes Q\in\bR^{nc\times nc}$ be the \emph{Kronecker product} of $P$ and $Q$. Due to the symmetry and the positive definiteness of $P$ and $Q$, the Kronecker product $P\otimes Q$ is also symmetric and positive definite, and $\cH_{P,Q}$ in \eqref{eq:hypo-spa-mul} could be defined equivalently as
\begin{equation}
\label{eq:hypo-spa-mul-equiv}%
\cH_{P,Q} :=\{H\mid\vect(H)^\T(P\otimes Q)\vect(H)\le1\}.
\end{equation}
The equivalence of Eqs.~\eqref{eq:hypo-spa-mul} and \eqref{eq:hypo-spa-mul-equiv} comes from the fact that $\tr(H^\T QHP)=\vect(H)^\T(P\otimes Q)\vect(H)$ following the well-known identity (see, e.g., Theorem~13.26 of \citealp{laub05})
\[ (P^\T\otimes Q)\vect(H)=\vect(QHP). \]
As a consequence, there is a bijection between $\cH_{P,Q}$ and
\[ \cE_{P,Q} :=\{\vh\mid\vh^\T(P\otimes Q)\vh\le1\} \]
which is geometrically an origin-centered ellipsoid in $\bR^{nc}$. The set of sign vectors $\{\sign(\vh)\mid\vh\in\cE_{P,Q}\}$ spreads over all the $N=2^{nc}$ quadrants of $\bR^{nc}$, and thus the set of sign matrices $\{\sign(H)\mid H\in\cH_{P,Q}\}$
contains all of $N$ possible dichotomies of $X_n\times\{1,\ldots,c\}$. In other words, $\cH_{P,Q}$ can be partitioned into $N$ equivalence classes $\cC_1,\ldots,\cC_N$, such that for fixed $k$, all soft response matrices in $\cC_k$ will generate the same labeling of $X_n\times\{1,\ldots,c\}$.

The definition of the power is same as before, and so is the definition of the volume:
\[ \cV(\cC_k):=\int_{\cC_k}\mathrm{d}H,\quad k=1,\ldots,N. \]
Because of the bijection between $\cH_{P,Q}$ and $\cE_{P,Q}$, $\cV(\cC_k)$ is likewise the geometric volume of all stacked soft response vectors in the intersection of the $k$-th quadrant of $\bR^{nc}$ and $\cE_{P,Q}$. By a similar argument to the definition of $V(\vh)$, we define
\begin{equation}
\label{eq:volume-mul}%
V(H):=\frac{\vh^\T(P\otimes Q)\vh}{\|\vh\|_2^2}=\frac{\tr(H^\T QHP)}{\|H\|_\mathrm{Fro}^2},
\end{equation}
where $\vh=\vect(H)$ and $\|H\|_\mathrm{Fro}$ means the Frobenius norm of $H$. We subsequently expect $V(H)$ to be small when $H$ lies in a large-volume equivalence class, and conversely to be large when $H$ lies in a small-volume equivalence class.

Note that $V(H)$ and $V(\vh)$ are consistent for binary learning problem settings. We can constrain $\vh_1+\vh_2=\zero_n$ if $c=2$ where $\zero_n$ is the all-zero vector in $\bR^n$. Let $P=I_2$ where $I_2$ is the identity matrix of size 2, then
\[ V(H) =\frac{\vh_1^\T Q\vh_1+\vh_2^\T Q\vh_2}{\|\vh_1\|_2^2+\|\vh_2\|_2^2}
=\frac{\vh_1^\T Q\vh_1}{\|\vh_1\|_2^2}=V(\vh_1), \]
which coincides with $V(\vh)$ defined in Eq.~\eqref{eq:volume-bin}. Similarly to $V(\vh)$, for two soft response matrices $H$ and $H'$ from the same equivalence class, $V(H)$ and $V(H')$ may not necessarily be the same value. In addition, the domain of $V(H)$ could be extended to $\bR^{n\times c}$ though the definition of $V(H)$ is originally null for $H$ outside $\cH_{P,Q}$.

\section{Multi-class Approximate Volume Regularization}

The proposed volume approximation motivates a family of new transductive methods taking it as a regularization. We develop and analyze an instantiation in this section whose optimization problem is non-convex but can be solved exactly and efficiently.

\subsection{Model}

First of all, we define the label indicator matrix $Y\in\bR^{n\times c}$ for convenience whose entries can be from either $\{0,1\}$ or $\{-1,0,1\}$ depending on the problem settings and whether negative labels ever appear. Specifically, we can set $Y_{i,j}=1$ if $x_i$ is labeled to have the $j$-th label and $Y_{i,j}=0$ otherwise, or alternatively we can set $Y_{i,j}=1$ if $x_i$ is labeled to have the $j$-th label, $Y_{i,j}=-1$ if $x_i$ is labeled to not have the $j$-th label, and $Y_{i,j}=0$ otherwise.

Let $\Delta(Y,H)$ be our loss function measuring the difference between $Y$ and $H$. The multi-class volume approximation motivates the following family of transductive methods:
\[ \min_{H\in\cH_{P,Q}} \Delta(Y,H)+\gamma\cdot\frac{\tr(H^\T QHP)}{\|H\|_\mathrm{Fro}^2}, \]
where $\gamma>0$ is a regularization parameter. The denominator $\|H\|_\mathrm{Fro}^2$ is quite difficult to tackle, so we would like to eliminate it as \citet{elyaniv08} and \citet{niu13}. We fix a scale parameter $\tau>0$, constrain $H$ to be of norm $\tau$, replace the feasible region $\cH_{P,Q}$ with $\bR^{n\times c}$ by extending the domain of $V(H)$ implicitly, and it becomes
\begin{equation}
\label{eq:mavr-proto}%
\begin{split}
\min_{H\in\bR^{n\times c}} &\;\Delta(Y,H)+\gamma\tr(H^\T QHP)\\
\ST \;\;&\; \|H\|_\mathrm{Fro}=\tau.
\end{split}
\end{equation}
Although the optimization in \eqref{eq:mavr-proto} is done in $\bR^{n\times c}$, the regularization is carried out relative to $\cH_{P,Q}$, since under the constraint $\|H\|_\mathrm{Fro}=\tau$, the regularization $\tr(H^\T QHP)$ is a weighted sum of the squares of cosines between $\vect(H)$ and the principal axes of $\cE_{P,Q}$ like \citet{elyaniv08}.

Subsequently, we denote by $\vy_1,\ldots,\vy_n$ and $\vr_1,\ldots,\vr_n$ the $c$-dimensional vectors that satisfy $Y=(\vy_1,\ldots,\vy_n)^\T$ and $H=(\vr_1,\ldots,\vr_n)^\T$. Consider the following loss functions to be $\Delta(Y,H)$ in optimization \eqref{eq:mavr-proto}:
\begin{enumerate}
  \setlength{\itemsep}{0em}%
  \item Squared losses over all data $\sum_{X_n}\|\vy_i-\vr_i\|_2^2$;
  \item Squared losses over labeled data $\sum_{X_l}\|\vy_i-\vr_i\|_2^2$;
  \item Linear losses over all data $\sum_{X_n}-\vy_i^\T\vr_i$;
  \item Linear losses over labeled data $\sum_{X_l}-\vy_i^\T\vr_i$;
\end{enumerate}
The first loss function has been used for multi-class transductive learning \citep{zhou03} and the binary counterparts of the fourth and third loss functions have been used for binary transductive learning \citep{elyaniv08} and clustering \citep{niu13}. Actually, the third and fourth loss functions are identical since $\vy_i$ for $x_i\in X_u$ is identically zero, and the first loss function is equivalent to them in \eqref{eq:mavr-proto} since $\sum_{X_n}\|\vy_i\|_2^2$ and $\sum_{X_l}\|\vy_i\|_2^2$ are constants and $\sum_{X_n}\|\vr_i\|_2^2=\tau^2$ is also a constant. The second loss function is undesirable for \eqref{eq:mavr-proto} due to an issue of the time complexity which will be discussed later. Thus, we instantiate $\Delta(Y,H):=\sum_{X_n}\|\vy_i-\vr_i\|_2^2=\|Y-H\|_\mathrm{Fro}^2$, and optimization \eqref{eq:mavr-proto} becomes
\begin{equation}
\label{eq:mavr}%
\begin{split}
\min_{H\in\bR^{n\times c}} &\;\|Y-H\|_\mathrm{Fro}^2+\gamma\tr(H^\T QHP)\\
\ST \;\;&\; \|H\|_\mathrm{Fro}=\tau.
\end{split}
\end{equation}
We refer to constrained optimization problem \eqref{eq:mavr} as \emph{multi-class approximate volume regularization} (MAVR). An unconstrained version of MAVR is then
\begin{equation}
\label{eq:mavr-uncon}%
\min_{H\in\bR^{n\times c}} \|Y-H\|_\mathrm{Fro}^2+\gamma\tr(H^\T QHP).
\end{equation}

\subsection{Algorithm}

Optimization \eqref{eq:mavr} is non-convex, but we can rewrite it using the stacked soft response vector $\vh=\vect(H)$ as
\begin{equation}
\label{eq:mavr-equiv}%
\begin{split}
\min_{\vh\in\bR^{nc}} &\;\|\vy-\vh\|_2^2+\gamma\vh^\T(P\otimes Q)\vh\\
\ST \;\;&\; \|\vh\|_2=\tau,
\end{split}
\end{equation}
where $\vy=\vect(Y)$ is the vectorization of $Y$. In this representation, the objective is a second-degree polynomial and the constraint is an origin-centered sphere, and fortunately we could solve it exactly and efficiently following \citet{forsythe65}. To this end, a fundamental property of the Kronecker product is necessary (see, e.g., Theorems~13.10 and 13.12 of \citealp{laub05}):

\begin{theorem}
\label{thm:kron-eigen}%
Let $\lambda_{Q,1}\le\cdots\le\lambda_{Q,n}$ be the eigenvalues and $\vv_{Q,1},\ldots,\vv_{Q,n}$ be the associated orthonormal eigenvectors of $Q$, $\lambda_{P,1}\le\cdots\le\lambda_{P,c}$ and $\vv_{P,1},\ldots,\vv_{P,c}$ be those of $P$, and the eigen-decompositions of $Q$ and $P$ be $Q=V_Q\Lambda_QV_Q^\T$ and $P=V_P\Lambda_PV_P^\T$. Then, the eigenvalues of $P\otimes Q$ are $\lambda_{P,j}\lambda_{Q,i}$ associated with orthonormal eigenvectors $\vv_{P,j}\otimes\vv_{Q,i}$ for $j=1,\ldots,c$, $i=1,\ldots,n$, and the eigen-decomposition of $P\otimes Q$ is $P\otimes Q=V_{PQ}\Lambda_{PQ}V_{PQ}^\T$, where $\Lambda_{PQ}=\Lambda_P\otimes\Lambda_Q$ and $V_{PQ}=V_P\otimes V_Q$.
\end{theorem}

After we ignore the constants $\|\vy\|_2^2$ and $\|\vh\|_2^2$ in the objective of optimization \eqref{eq:mavr-equiv}, the Lagrange function is
\[ \Phi(\vh,\rho)=-2\vh^\T\vy+\gamma\vh^\T(P\otimes Q)\vh-\rho(\vh^\T\vh-\tau^2), \]
where $\rho\in\bR$ is the Lagrangian multiplier for $\|\vh\|_2^2=\tau^2$. The stationary conditions are
\begin{align}
\label{eq:stat-cond1}%
\partial\Phi/\partial\vh &= -\vy+\gamma(P\otimes Q)\vh-\rho\vh = \zero_{nc},\\
\label{eq:stat-cond2}%
\partial\Phi/\partial\rho &= \vh^\T\vh-\tau^2 = 0.
\end{align}
Hence, for any locally optimal solution $(\vh,\rho)$ where $\rho/\gamma$ is not an eigenvalue of $P\otimes Q$, we have
\begin{align}
\label{eq:opt-h}%
\vh&=(\gamma P\otimes Q-\rho I_{nc})^{-1}\vy\\
&=V_{PQ}(\gamma\Lambda_{PQ}-\rho I_{nc})^{-1}V_{PQ}^\T\vy\notag\\
\label{eq:opt-h-eig}%
&=(V_P\otimes V_Q)(\gamma\Lambda_{PQ}-\rho I_{nc})^{-1}\vect(V_Q^\T YV_P)
\end{align}
based on Eq.~\eqref{eq:stat-cond1} and Theorem~\ref{thm:kron-eigen}. Next, we search for the feasible $\rho$ for \eqref{eq:stat-cond1} and \eqref{eq:stat-cond2} which will lead to the globally optimal $\vh$. Let $\vz=\vect(V_Q^\T YV_P)$, then plugging \eqref{eq:opt-h-eig} into \eqref{eq:stat-cond2} gives us
\begin{equation}
\label{eq:stat-cond3}%
\vz^\T(\gamma\Lambda_{PQ}-\rho I_{nc})^{-2}\vz-\tau^2=0.
\end{equation}
Let us sort the eigenvalues $\lambda_{P,1}\lambda_{Q,1},\ldots,\lambda_{P,c}\lambda_{Q,n}$ into a non-descending sequence $\{\lambda_{PQ,1},\ldots,\lambda_{PQ,nc}\}$, rearrange $\{z_1,\ldots,z_{nc}\}$ accordingly, and find the smallest $k_0$ which satisfies $z_{k_0}\neq0$. As a result, Eq.~\eqref{eq:stat-cond3} implies that
\begin{equation}
\label{eq:grho-gen}%
g(\rho)=\sum_{k=k_0}^{nc}\frac{z_k^2}{(\gamma\lambda_{PQ,k}-\rho)^2}-\tau^2=0
\end{equation}
for any stationary $\rho$. By Theorem~4.1 of \citet{forsythe65}, the smallest root of $g(\rho)$ determines a unique $\vh$ so that $(\vh,\rho)$ is the globally optimal solution to $\Phi(\vh,\rho)$, i.e., $\vh$ minimizes the objective of \eqref{eq:mavr-equiv} globally. For this $\rho$, the only exception when it cannot determine $\vh$ by Eq.~\eqref{eq:opt-h} is that $\rho/\gamma$ is an eigenvalue of $P\otimes Q$, but this happens with probability zero. Finally, the theorem below points out the location of this $\rho$ (the proof is in the appendix):

\begin{theorem}
\label{thm:opt-rho}%
The function $g(\rho)$ defined in Eq.~\eqref{eq:grho-gen} has exactly one root in the interval $[\rho_0,\gamma\lambda_{PQ,k_0})$ and no root in the interval $(-\infty,\rho_0)$, where $\rho_0=\gamma\lambda_{PQ,k_0}-\|\vy\|_2/\tau$.
\end{theorem}

\begin{algorithm}[t]
\caption{MAVR}
\label{alg:mavr}%
\begin{algorithmic}
    \STATE \textbf{Input:} $P$, $Q$, $Y$, $\gamma$ and $\tau$
    \STATE \textbf{Output:} $H$ and $\rho$
\end{algorithmic}
\begin{algorithmic}[1]
    \STATE Eigen-decompose $P$ and $Q$;
    \STATE Construct the function $g(\rho)$;
    \STATE Find the smallest root of $g(\rho)$;
    \STATE Recover $\vh$ using $\rho$ and reshape $\vh$ to $H$.
\end{algorithmic}
\end{algorithm}

The algorithm of MAVR is summarized in Algorithm~\ref{alg:mavr}. It is easy to see that fixing $\rho=-1$ in Algorithm~\ref{alg:mavr} instead of finding the smallest root of $g(\rho)$ suffices to solve optimization \eqref{eq:mavr-uncon}. Moreover, for a special case $P=I_c$ where $I_c$ is the identity matrix of size $c$, any stationary $H$ is simply
\[ H=(\gamma Q-\rho I_n)^{-1}Y=V_Q(\gamma\Lambda_Q-\rho I_n)^{-1}V_Q^\T Y. \]
Let $\vz=V_Q^\T Y\one_c$ where $\one_c$ is the all-one vector in $\bR^c$, and $k_0$ is the smallest number that satisfies $z_{k_0}\neq0$. Then the smallest root of
\[\textstyle g(\rho)=\sum_{k=k_0}^nz_k^2/(\gamma\lambda_{Q,k}-\rho)^2-\tau^2 \]
gives us the feasible $\rho$ leading to the globally optimal $H$.

The asymptotic time complexity of Algorithm~\ref{alg:mavr} is $O(n^3)$. More specifically, eigen-decomposing $Q$ in the first step of Algorithm~\ref{alg:mavr} costs $O(n^3)$, and this is the dominating computation time. Eigen-decomposing $P$ just needs $O(c^3)$ and is negligible under the assumption that $n\gg c$ without loss of generality. In the second step, it requires $O(nc\log(nc))$ for sorting the eigenvalues of $P\otimes Q$ and $O(n^2c)$ for computing $\vz$. Finding the smallest root of $g(\rho)$ based on a binary search algorithm uses $O(\log(\|\vy\|_2))$ in the third step, and $\|\vy\|_2\le\sqrt{l}$ for multi-class problems and $\|\vy\|_2\le\sqrt{lc}$ for multi-label problems. In the final step, recovering $\vh$ is essentially same as computing $\vz$ and costs $O(n^2c)$.

We would like to comment a bit more on the asymptotic time complexity of MAVR. Firstly, we employ the squared losses over all data rather than the squared losses over labeled data. If the latter loss function was plugged in optimization \eqref{eq:mavr-proto}, Eq.~\eqref{eq:opt-h} would become
\[ \vh=(\gamma P\otimes Q-\rho I_{nc}+I_c\otimes J)^{-1}\vy, \]
where $J$ is an $n$-by-$n$ diagonal matrix such that $J_{i,i}=1$ if $x_i$ is labeled and $J_{i,i}=0$ if $x_i$ is unlabeled. The inverse in the expression above cannot be computed using the eigen-decompositions of $P$ and $Q$, and hence the computational complexity would increase from $O(n^3)$ to $O(n^3c^3)$. Secondly, given fixed $P$ and $Q$ but different $Y$, $\gamma$, and $\tau$, the computational complexity is $O(n^2c)$ if we reuse the eigen-decompositions of $P$ and $Q$ and the sorted eigenvalues of $P\otimes Q$. This property is especially advantageous for validating and selecting hyperparameters. It is also quite useful for picking different $X_l\subset X_n$ to be labeled following transductive problem settings. Finally, the asymptotic time complexity $O(n^3)$ can hardly be improved based on existing techniques for optimizations \eqref{eq:mavr} and \eqref{eq:mavr-uncon}. Even if $\rho$ is fixed in optimization \eqref{eq:mavr-uncon}, the stationary condition Eq.~\eqref{eq:stat-cond1} is a \emph{discrete Sylvester equation} which consumes $O(n^3)$ for solving it \citep{sima96}.

\subsection{Theoretical analyses}

We provide two theoretical results. Under certain assumptions, the stability analysis upper bounds the difference of two optimal $H$ and $H'$ trained with two different label indicator matrices $Y$ and $Y'$, and the error analysis bounds the difference of $H$ from the ground truth.

Theorem~\ref{thm:opt-rho} guarantees that $\rho<\gamma\lambda_{PQ,k_0}$. In fact, with high probability over the choice of $Y$, it holds that $k_0=1$ (we did not meet $k_0>1$ in our experiments). For this reason, we make the following assumption:\\[3pt]
\emph{Fix $P$ and $Q$, and allow $Y$ to change according to the partition of $X_n$ into different $X_l$ and $X_u$. There is $C_{\gamma,\tau}>0$, which just depends on $\gamma$ and $\tau$, such that for all optimal $\rho$ trained with different $Y$, $\rho\le\gamma\lambda_{PQ,1}-C_{\gamma,\tau}$.}

Note that for unconstrained MAVR, there must be $C_{\gamma,\tau}>1$ since $\gamma\lambda_{PQ,1}>0$ and $\rho=-1$. Based on the above assumption and the lower bound of $\rho$ in Theorem~\ref{thm:opt-rho}, we can prove the theorem below.

\begin{theorem}[Stability of MAVR]
\label{thm:stable}%
Assume the existence of $C_{\gamma,\tau}$. Let $(H,\rho)$ and $(H',\rho')$ be two globally optimal solutions trained with two different label indicator matrices $Y$ and $Y'$ respectively. Then,
\begin{equation}
\label{eq:stable}%
\|H-H'\|_\mathrm{Fro} \le \|Y-Y'\|_\mathrm{Fro}/C_{\gamma,\tau}
+|\rho-\rho'|\min\{\|Y\|_\mathrm{Fro},\|Y'\|_\mathrm{Fro}\}/C_{\gamma,\tau}^2.
\end{equation}
Consequently, for MAVR in optimization \eqref{eq:mavr} we have
\[
\|H-H'\|_\mathrm{Fro} \le \|Y-Y'\|_\mathrm{Fro}/C_{\gamma,\tau}
+\|Y\|_\mathrm{Fro}\|Y'\|_\mathrm{Fro}/\tau C_{\gamma,\tau}^2,
\]
and for unconstrained MAVR in optimization \eqref{eq:mavr-uncon} we have
\[ \|H-H'\|_\mathrm{Fro} \le \|Y-Y'\|_\mathrm{Fro}/C_{\gamma,\tau}. \]
\end{theorem}

In order to present an error analysis, we assume there is a ground-truth soft response matrix $H^*$ with two properties. Firstly, the value of $V(H^*)$ should be bounded, namely,
\[ V(H^*) = \frac{\tr(H^{*\T} QH^*P)}{\|H^*\|_\mathrm{Fro}^2} \le C_h, \]
where $C_h>0$ is a small number. This ensures that $H^*$ lies in a large-volume region. Otherwise MAVR implementing the large volume principle can by no means learn some $H$ close to $H^*$. Secondly, $Y$ should contain certain information about $H^*$.
MAVR makes use of $P$, $Q$ and $Y$ only and the meanings of $P$ and $Q$ are fixed already, so MAVR may access the information about $H^*$ only through $Y$. To make $Y$ and $H^*$ correlated, we assume that $Y=H^*+E$ where $E\in\bR^{n\times c}$ is a noise matrix of the same size as $Y$ and $H^*$. All entries of $E$ are independent with zero mean, and the variance of them is $\sigma_l$ or $\sigma_u$ depending on its correspondence to a labeled or an unlabeled position in $Y$. We could expect that $\sigma_l\ll\sigma_u$, such that the entries of $Y$ in labeled positions are close to the corresponding entries of $H^*$, but the entries of $Y$ in unlabeled positions are completely corrupted and uninformative for recovering $H^*$. Notice that we need this generating mechanism of $Y$ even if $C_h/\gamma$ is the smallest eigenvalue of $P\otimes Q$, since $P\otimes Q$ may have multiple smallest eigenvalues and $\pm H$ have totally different meanings. Based on these assumptions, we can prove the theorem below.

\begin{theorem}[Accuracy of MAVR]
\label{thm:error}%
Assume the existence of $C_{\gamma,\tau}$, $C_h$, and the generating process of $Y$ from $H^*$ and $E$. Let $\widetilde{l}$ and $\widetilde{u}$ be the numbers of the labeled and unlabeled positions in $Y$ and assume that $\bE_E\|Y\|_\mathrm{Fro}^2\le\widetilde{l}$ where the expectation is with respect to the noise matrix $E$. For each possible $Y$, let $H$ be the globally optimal solution trained with it. Then,
\begin{align}
& \bE_E\|H-H^*\|_\mathrm{Fro}
\le (\sqrt{C_h}\gamma\lambda_{PQ,1}/C_{\gamma,\tau})\|H^*\|_\mathrm{Fro}\notag\\
&\qquad +\left(\max\left\{\sqrt{\widetilde{l}}/\tau-\gamma\lambda_{PQ,1}-1,
\gamma\lambda_{PQ,1}-C_{\gamma,\tau}+1\right\}/C_{\gamma,\tau}\right)\|H^*\|_\mathrm{Fro}\notag\\
\label{eq:error-con}%
&\qquad +\sqrt{\widetilde{l}\sigma_l^2+\widetilde{u}\sigma_u^2}/C_{\gamma,\tau}
\end{align}
for MAVR in optimization \eqref{eq:mavr}, and
\begin{equation}
\label{eq:error-uncon}%
\bE_E\|H-H^*\|_\mathrm{Fro}^2 \le (C_h/4)\|H^*\|_\mathrm{Fro}^2
+\widetilde{l}\sigma_l^2+\widetilde{u}\sigma_u^2
\end{equation}
for unconstrained MAVR in optimization \eqref{eq:mavr-uncon}.
\end{theorem}

The proofs of Theorems~\ref{thm:stable} and \ref{thm:error} are in the appendix. Considering the instability bounds in Theorem~\ref{thm:stable} and the error bounds in Theorem~\ref{thm:error}, unconstrained MAVR is superior to constrained MAVR in both cases. That being said, bounds are just bounds. We will demonstrate the potential of constrained MAVR in the next section by experiments.

\section{Experiments}

In this section, we numerically evaluate MAVR.

\subsection{Serendipitous learning}

\begin{figure*}[t]
  \centering
  \subfigure[Data 1]{\includegraphics[width=0.24\columnwidth]{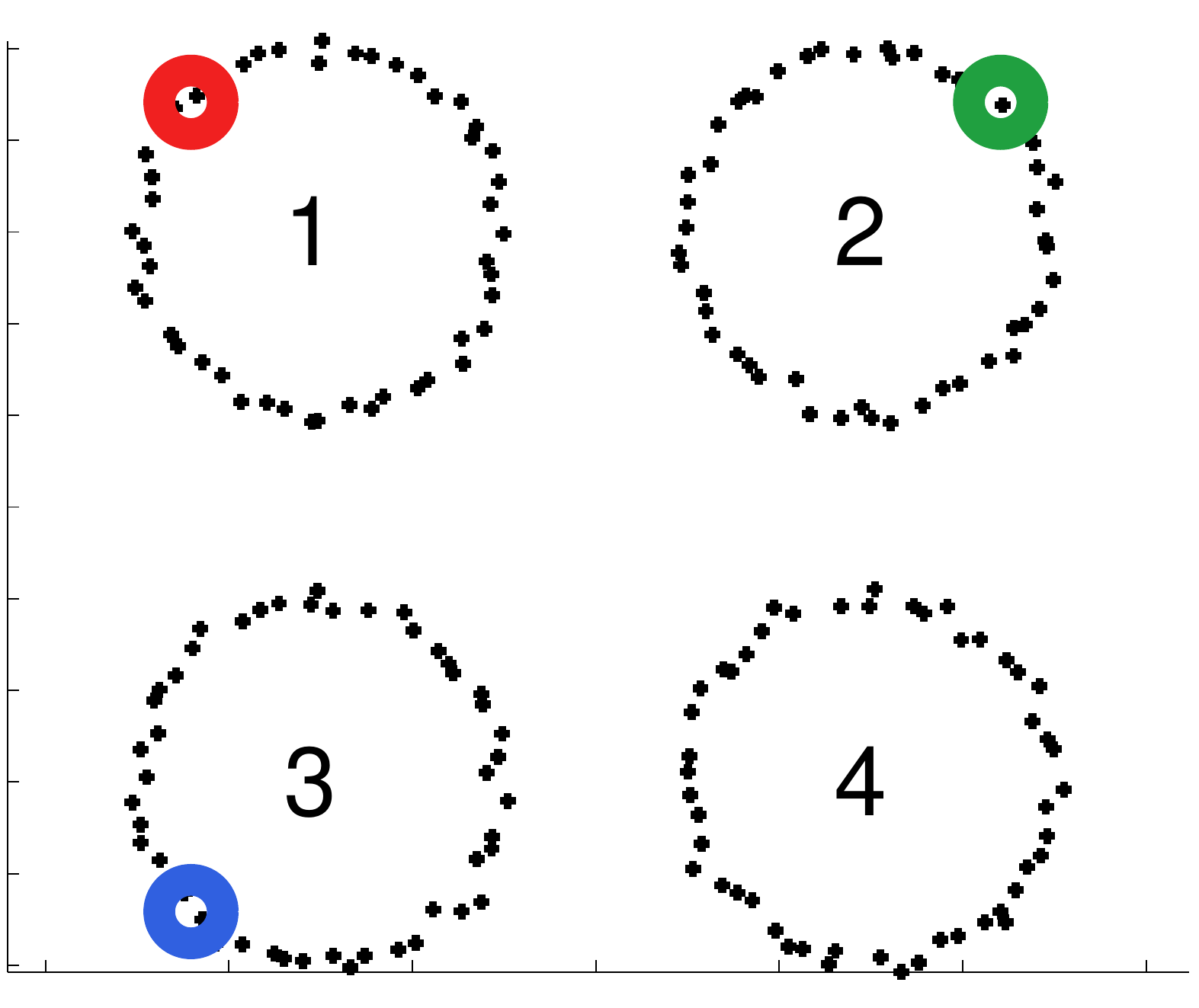}}
  \subfigure[Data 1, $P_1$]{\includegraphics[width=0.24\columnwidth]{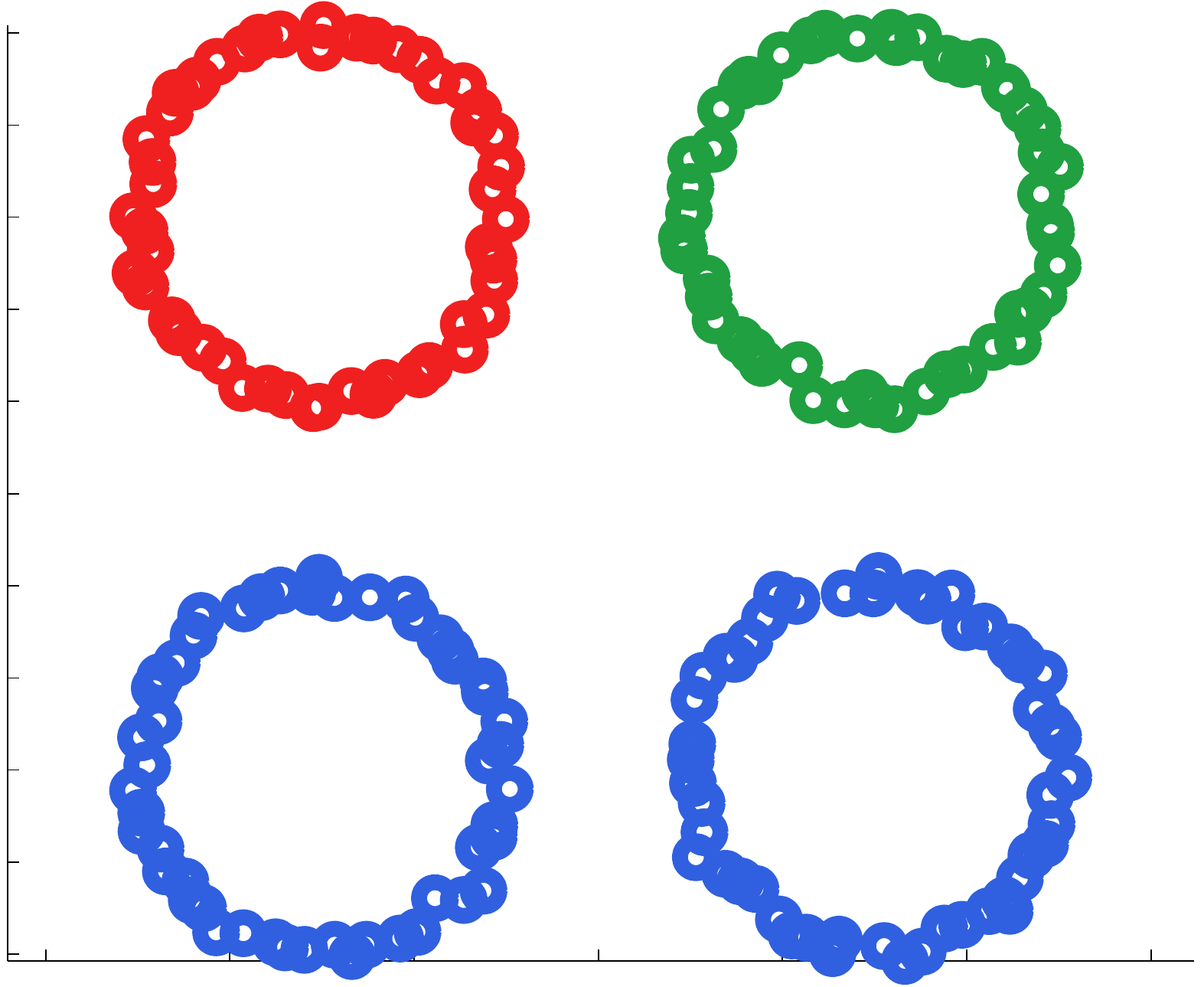}}
  \subfigure[Data 1, $P_2$]{\includegraphics[width=0.24\columnwidth]{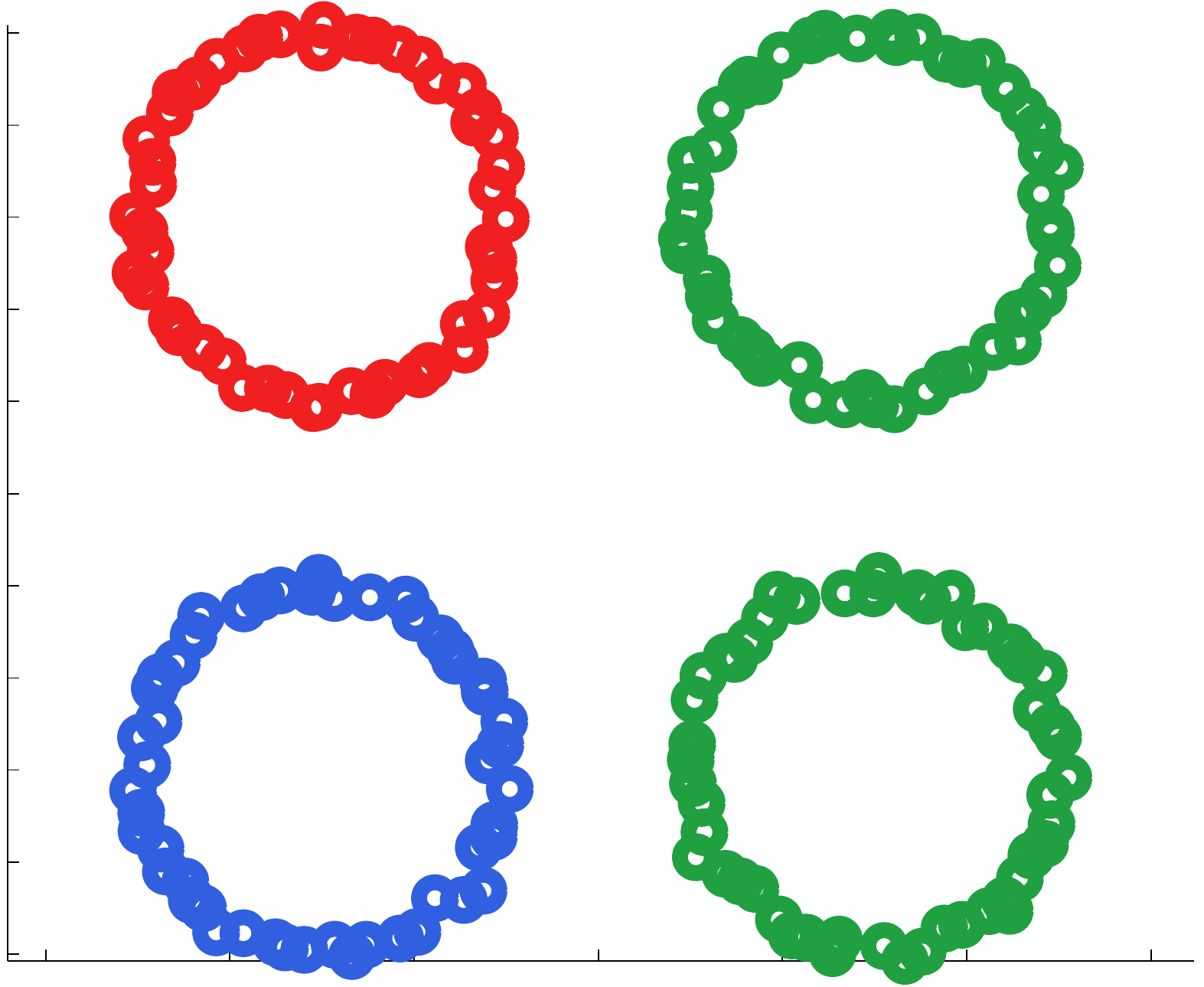}}
  \subfigure[Data 1, $P_3$]{\includegraphics[width=0.24\columnwidth]{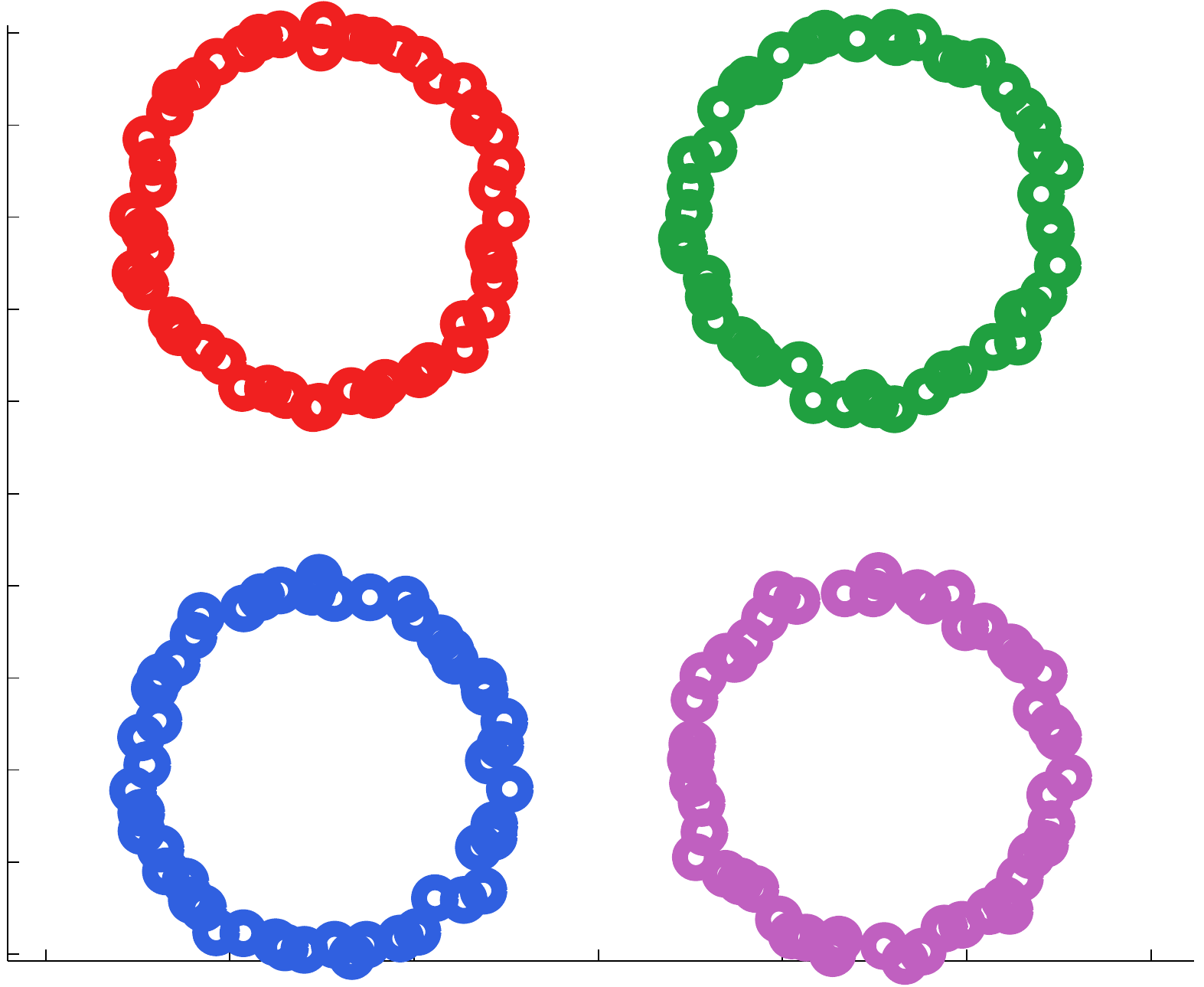}}\\
  \subfigure[Data 2]{\includegraphics[width=0.24\columnwidth]{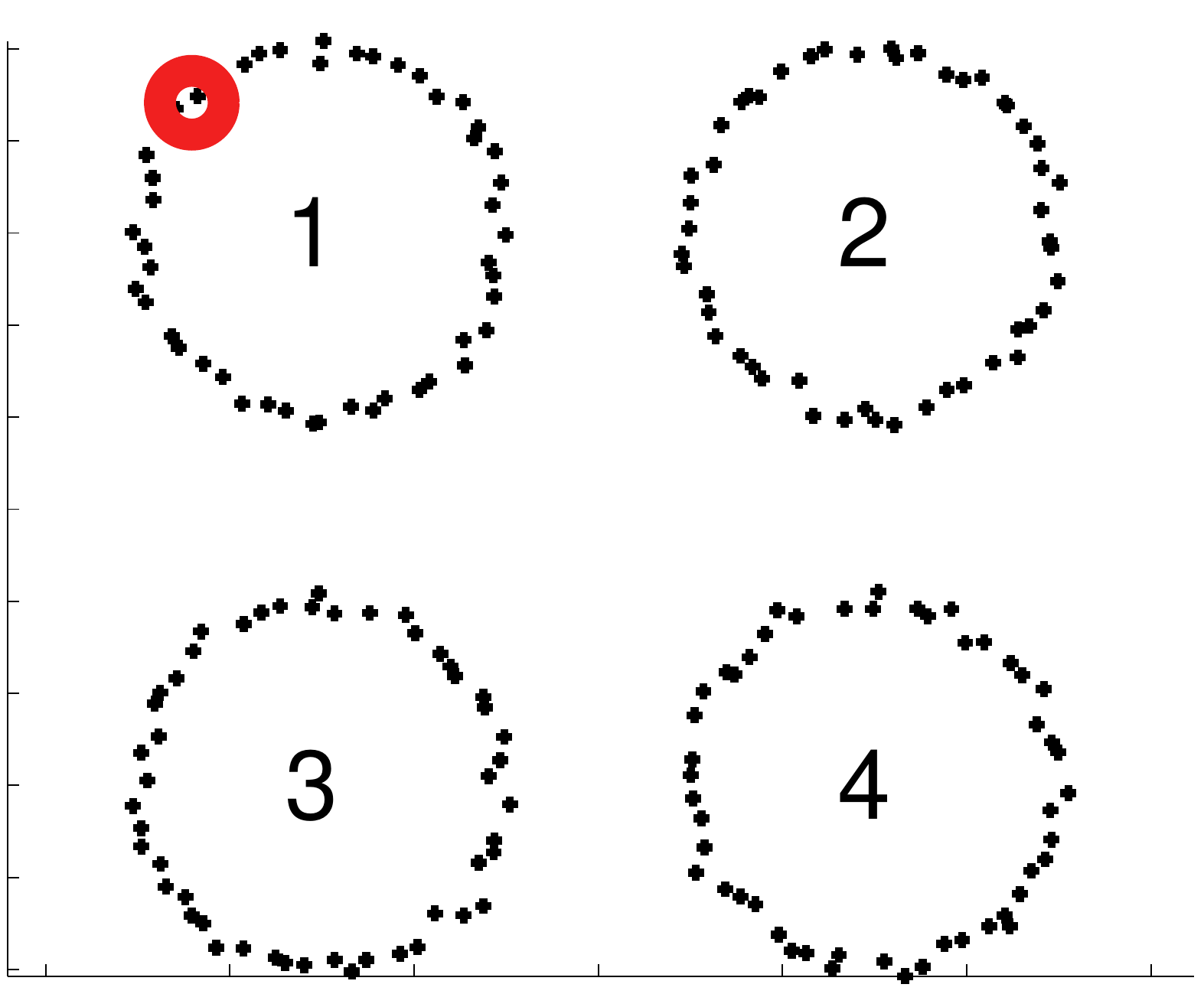}}
  \subfigure[Data 2, $P_4$]{\includegraphics[width=0.24\columnwidth]{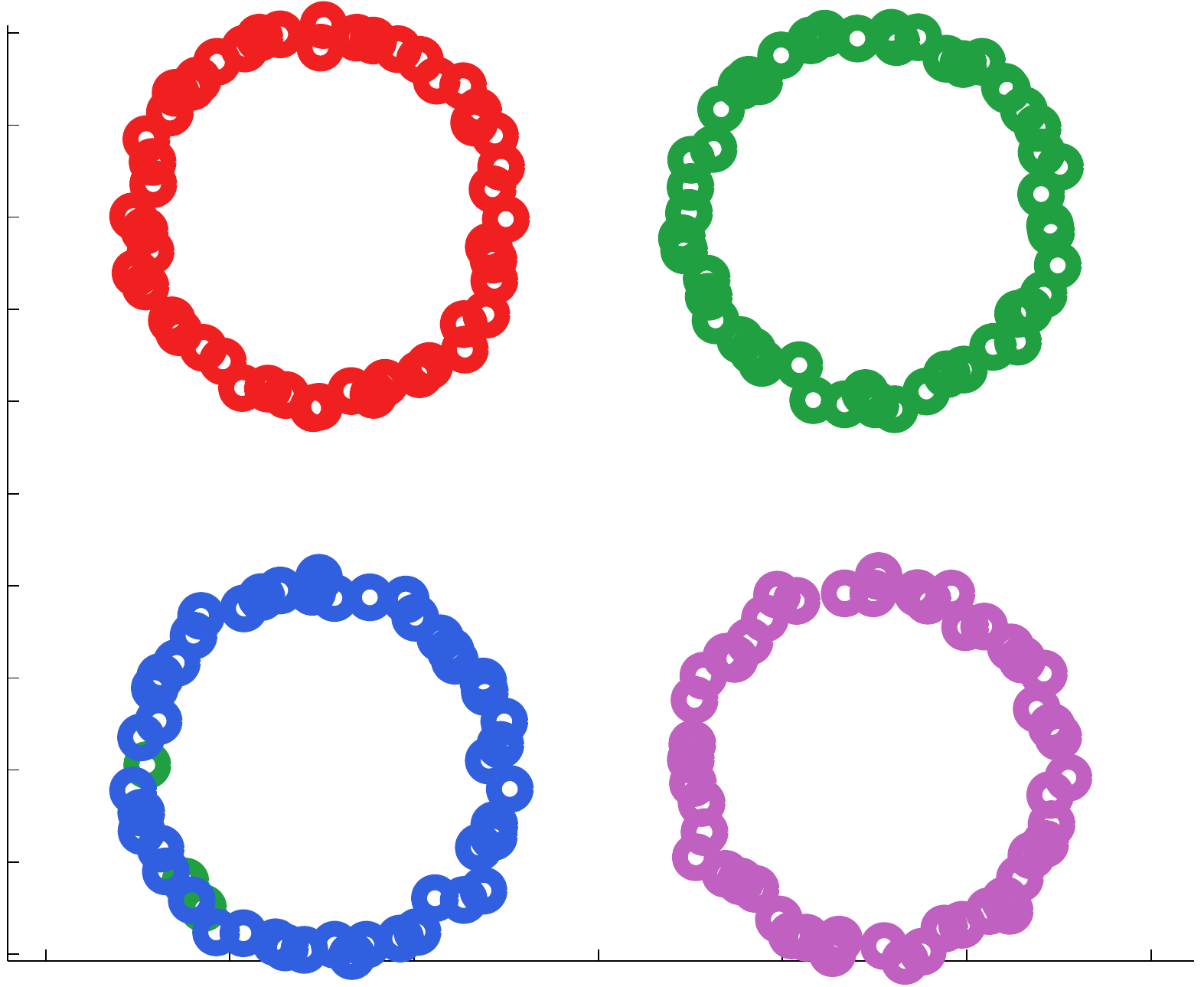}}
  \subfigure[Data 3]{\includegraphics[width=0.24\columnwidth]{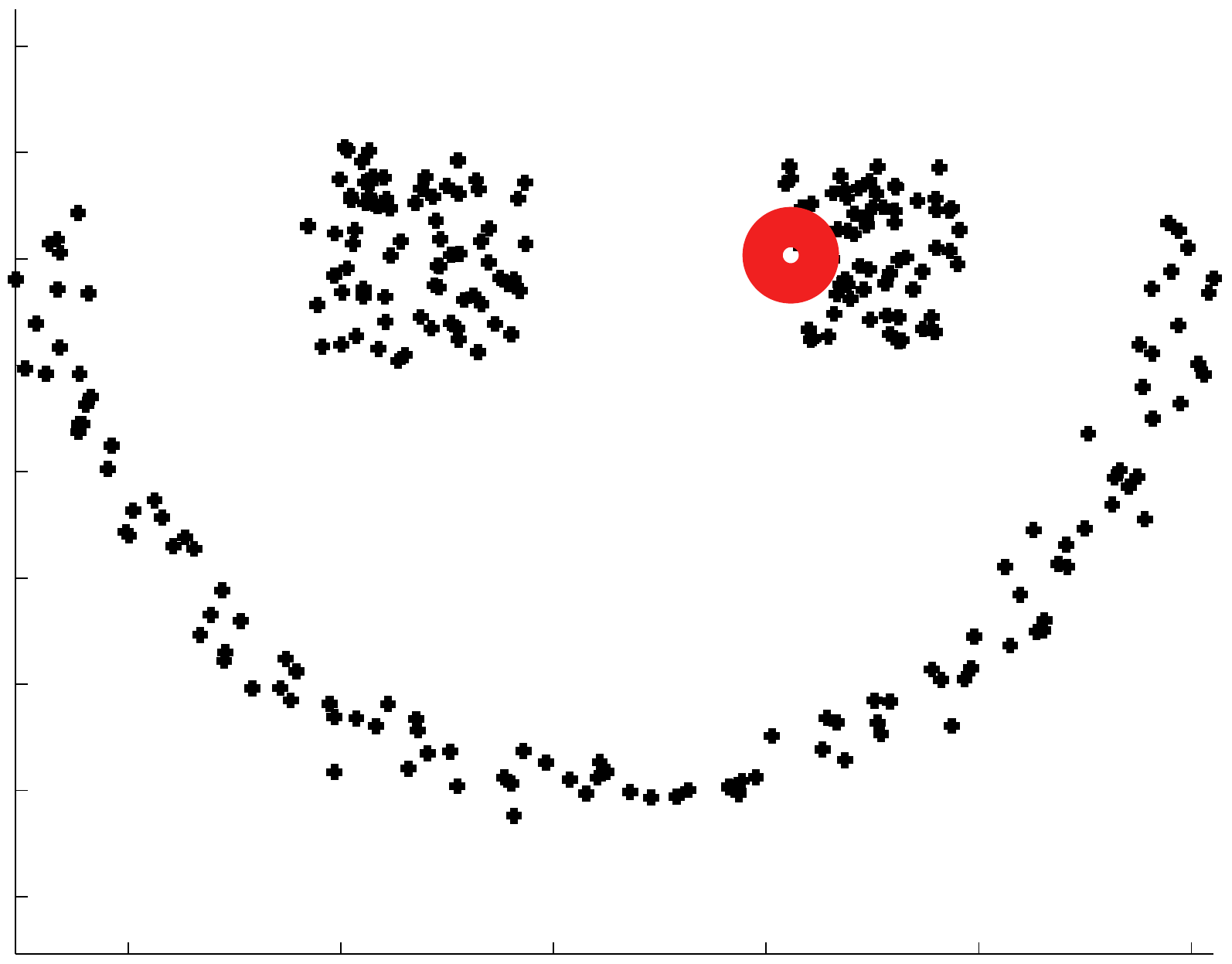}}
  \subfigure[Data 3, $P_5$]{\includegraphics[width=0.24\columnwidth]{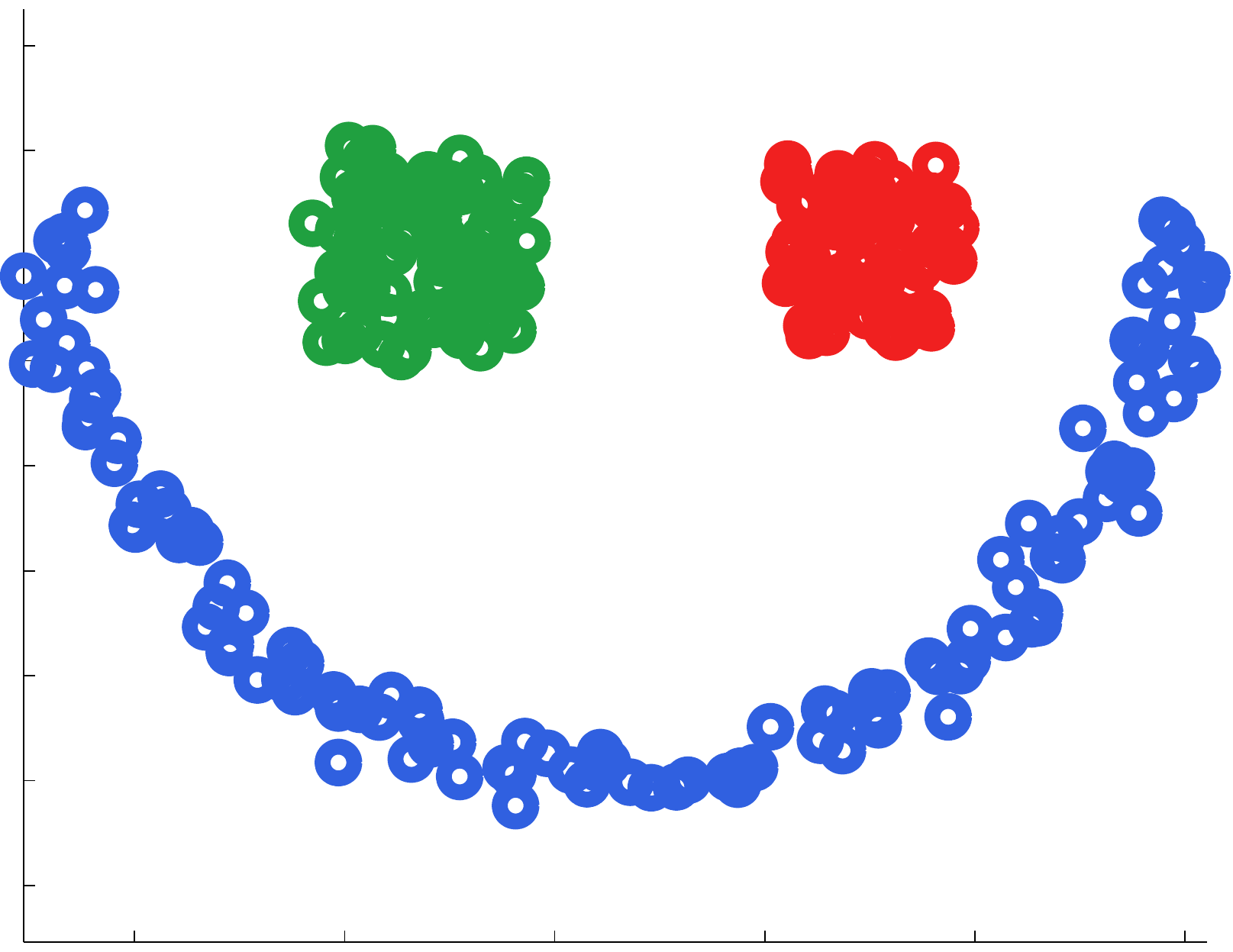}}\\
  \subfigure[Data 4]{\includegraphics[width=0.24\columnwidth]{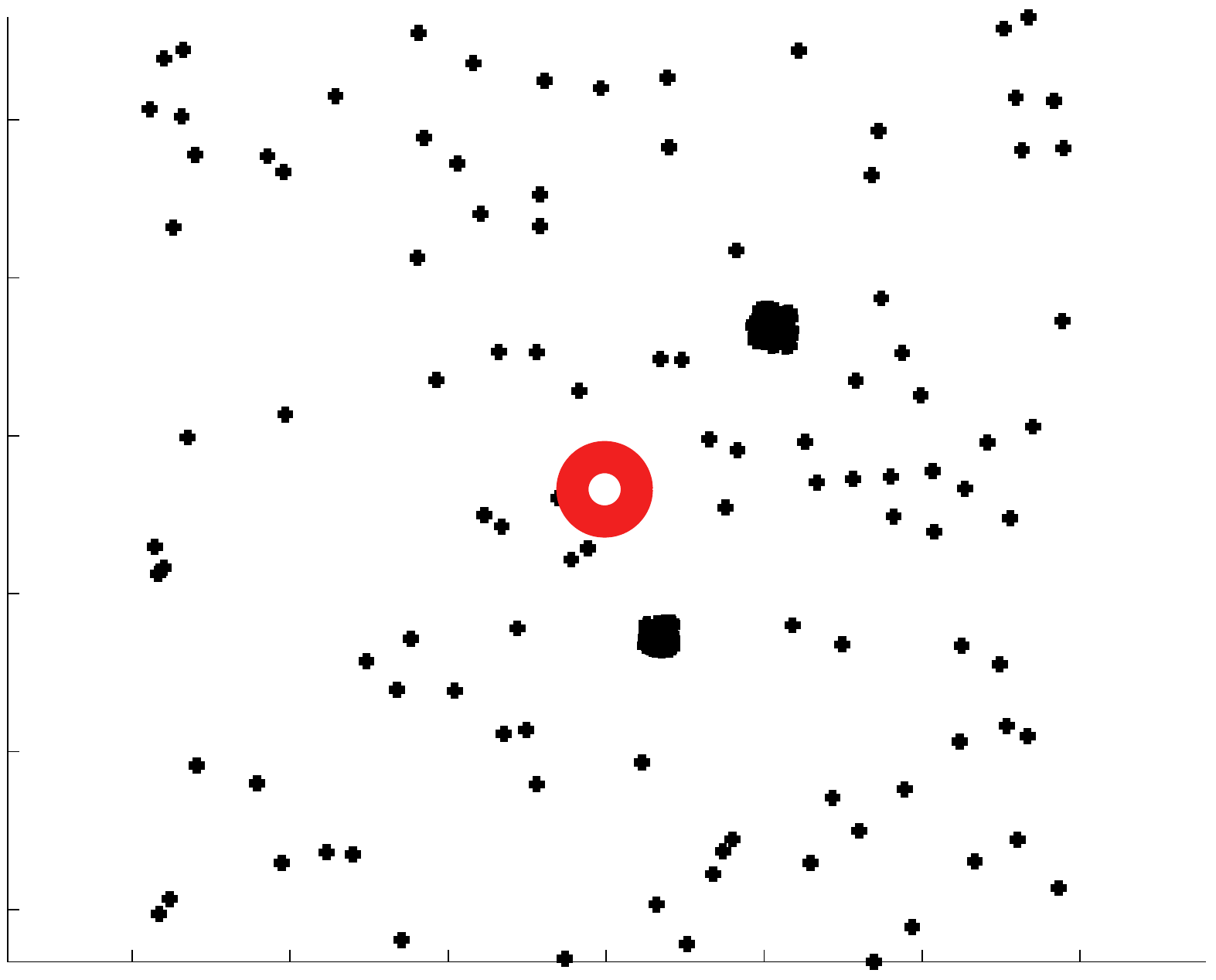}}
  \subfigure[Data 4, $P_5$]{\includegraphics[width=0.24\columnwidth]{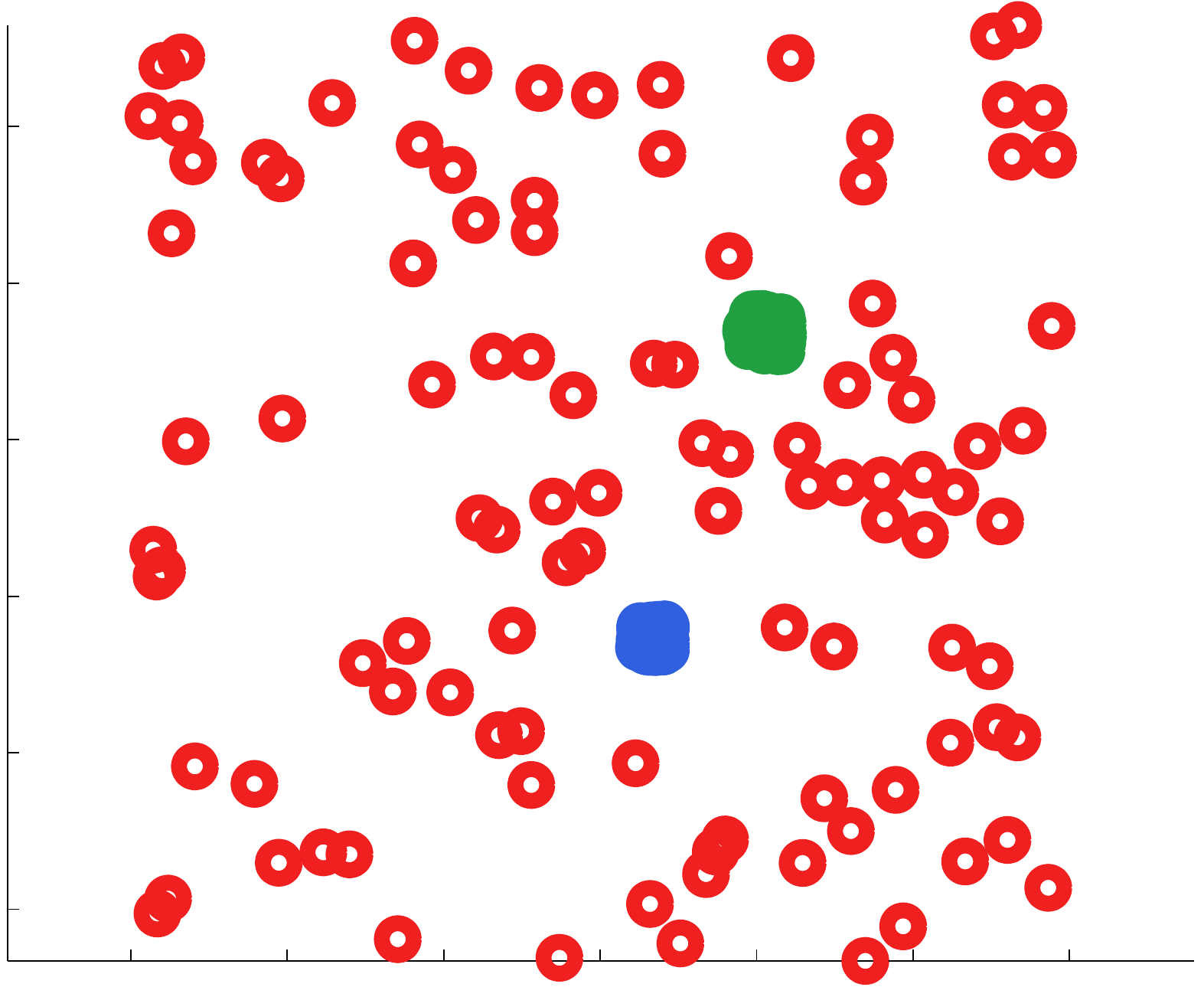}}
  \subfigure[Data 5]{\includegraphics[width=0.24\columnwidth]{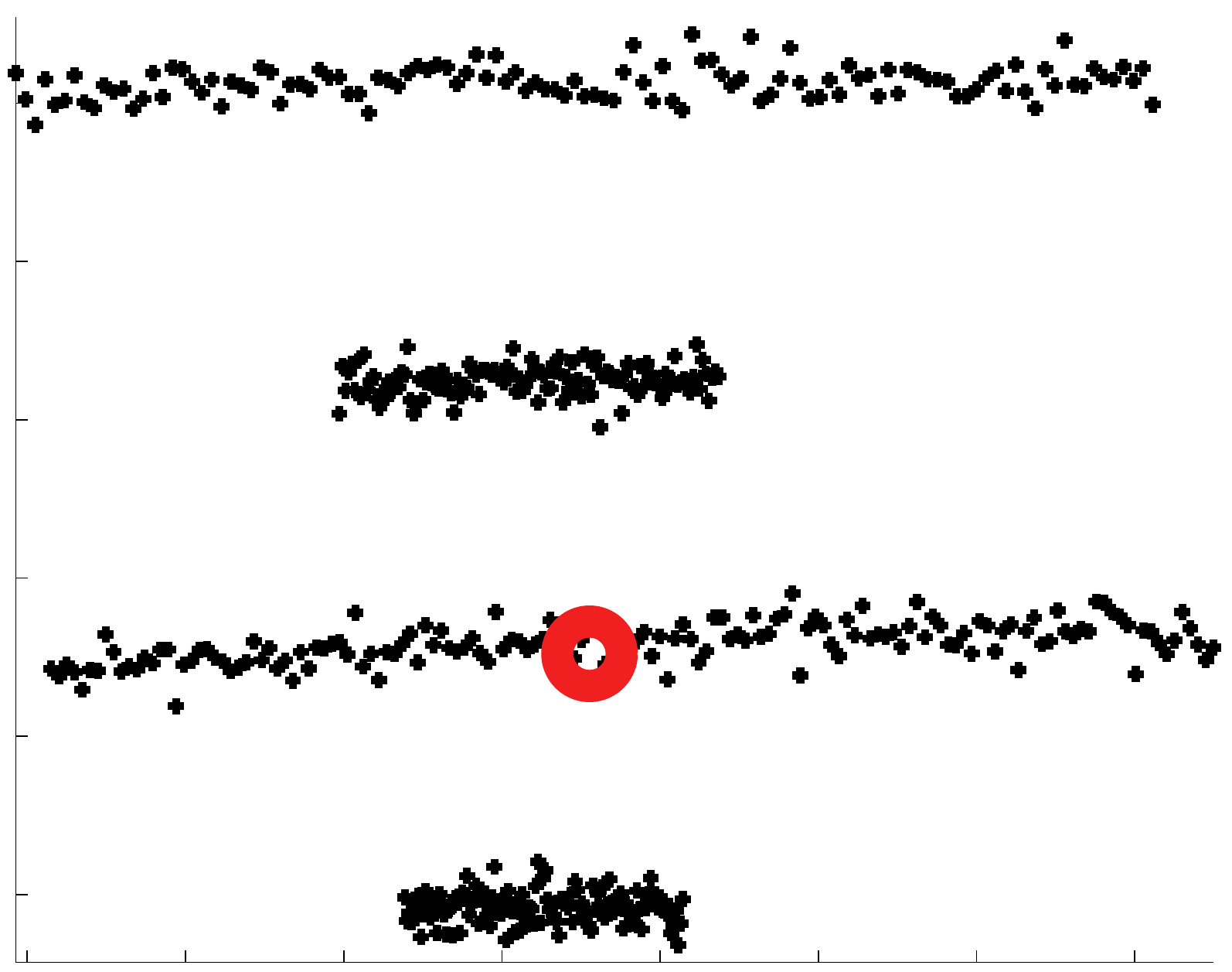}}
  \subfigure[Data 5, $P_6$]{\includegraphics[width=0.24\columnwidth]{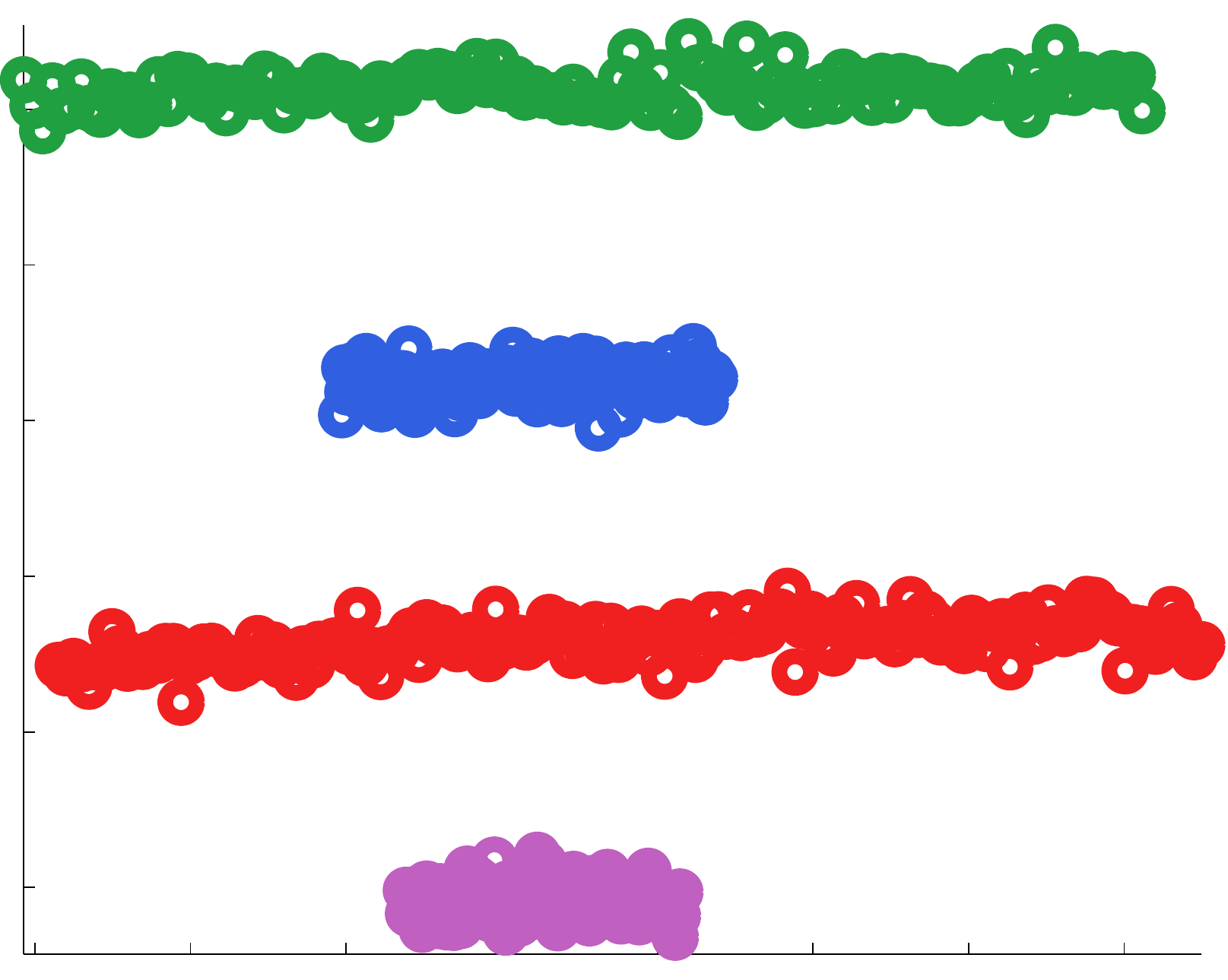}}
  \caption{Experimental results of serendipitous learning problems on artificial data sets.}
  \label{fig:exp-seren}
\end{figure*}

We show how to handle serendipitous problems by MAVR directly without performing clustering \citep{hartigan79,ng01,sugi14} or estimating the class-prior change \citep{christo12}. The experimental results are displayed in Figure~\ref{fig:exp-seren}. There are 5 data sets, and the latter 3 data sets are from \citet{zelnik04}. The matrix $Q$ was specified as the \emph{normalized graph Laplacian} (see, e.g., \citealp{luxburg07})\footnote{Though the graph Laplacian matrices have zero eigenvalues, they would not cause algorithmic problems when used as $Q$.} $L_\textrm{nor}=I_n-D^{-1/2}WD^{-1/2}$, where $W\in\bR^{n\times n}$ is a similarity matrix and $D\in\bR^{n\times n}$ is the degree matrix of $W$. The matrix $P$ was specified by
\begin{align*}
P_1 = \scalebox{1}{$\begin{pmatrix}1&0&0&0\\0&1&0&0\\0&0&3&1\\0&0&1&1\end{pmatrix}$},\;&
P_2 = \scalebox{1}{$\begin{pmatrix}1&0&0&0\\0&3&0&1\\0&0&1&0\\0&1&0&1\end{pmatrix}$},\;
P_3 = \scalebox{1}{$\begin{pmatrix}1&0&0&0\\0&1&0&1\\0&0&1&1\\0&1&1&3\end{pmatrix}$},\\
P_4 = \scalebox{1}{$\begin{pmatrix}1&1/2&1/2&1/2\\ 1/2&2&0&1/2\\ 1/2&0&2&1/2\\ 1/2&1/2&1/2&3\end{pmatrix}$},\;&
P_5 = \scalebox{1}{$\begin{pmatrix}1&1/2&1/2\\ 1/2&1&0\\ 1/2&0&1\end{pmatrix}$},\;
P_6 = \scalebox{1}{$\begin{pmatrix}1&1/2&1/2&1/2\\ 1/2&1&0&0\\ 1/2&0&1&0\\ 1/2&0&0&1\end{pmatrix}$}.
\end{align*}
For data sets 1 and 2 we used the Gaussian similarity
\[ W_{i,j}=\exp(-\|x_i-x_j\|_2^2/(2\sigma^2)) \]
with the kernel width $\sigma=0.25$, and for data sets 3 to 5 we applied the local-scaling similarity \citep{zelnik04}
\[ W_{i,j}=\exp(-\|x_i-x_j\|_2^2/(2\sigma_i\sigma_j)),\quad\sigma_i=\|x_i-x_i^{(k)}\|_2 \]
with the number of nearest neighbors $k=7$, where $x_i^{(k)}$ is the $k$-th nearest neighbor of $x_i$ in $X_n$. We set $\gamma=99$ and $\tau=\sqrt{l}$. Furthermore, a class balance regularization was imposed for data sets 2 to 5. The detail is omitted here due to the space limit, while the idea is to encourage balanced total responses of all $c$ classes. For this regularization, the regularization parameter was $\gamma'=1$. We can see that in Figure~\ref{fig:exp-seren}, MAVR successfully classified the data belonging to the known classes and simultaneously clustered the data belonging to the unknown classes. By specifying different $P$, we could control the influence of the known classes on the unknown classes.

\subsection{Multi-class learning}

\begin{figure*}[t]
  \centering
  \subfigure[Varying $\sigma_\epsilon$]{\includegraphics[width=0.3\columnwidth]{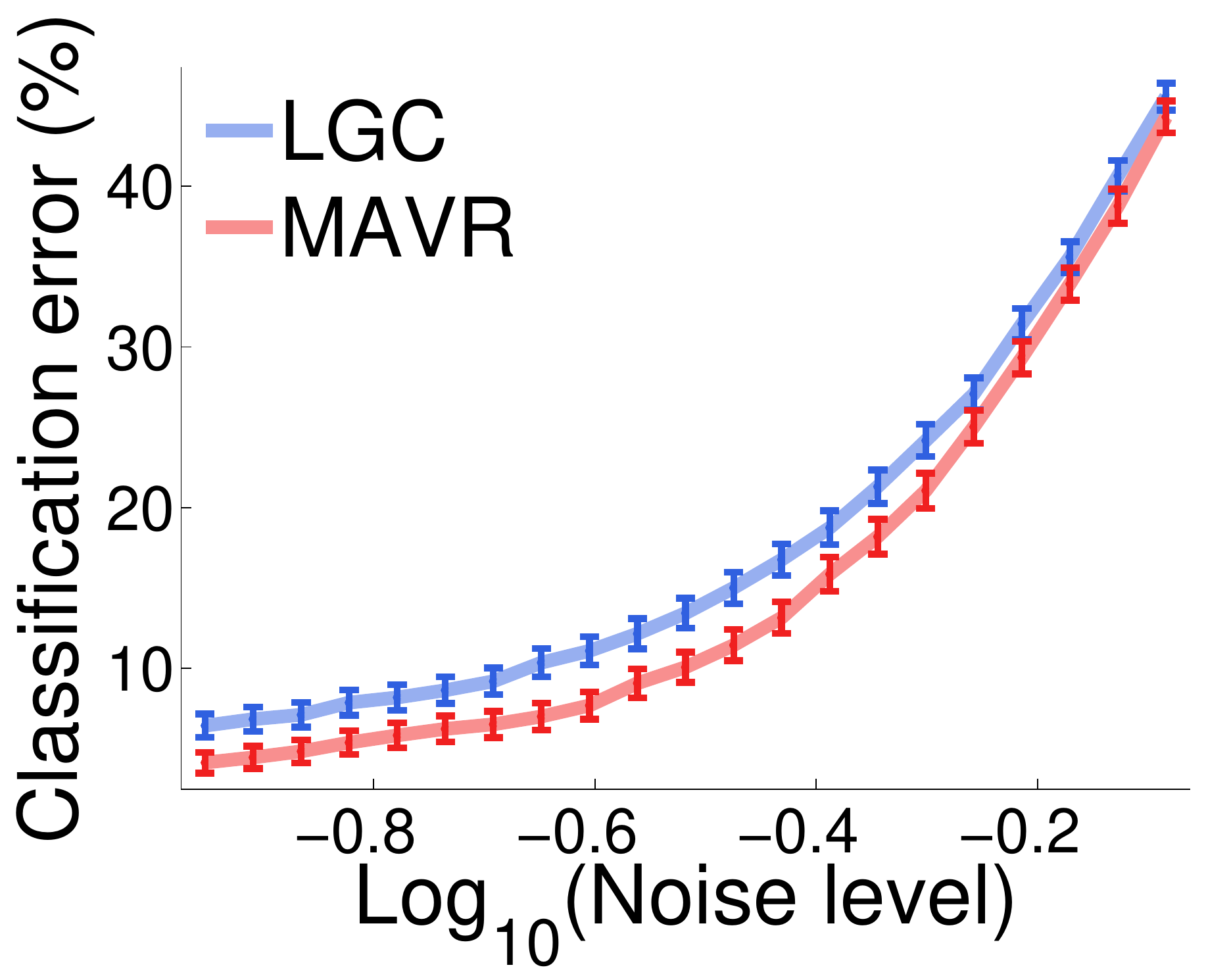}}~~~
  \subfigure[Varying $\sigma$]{\includegraphics[width=0.3\columnwidth]{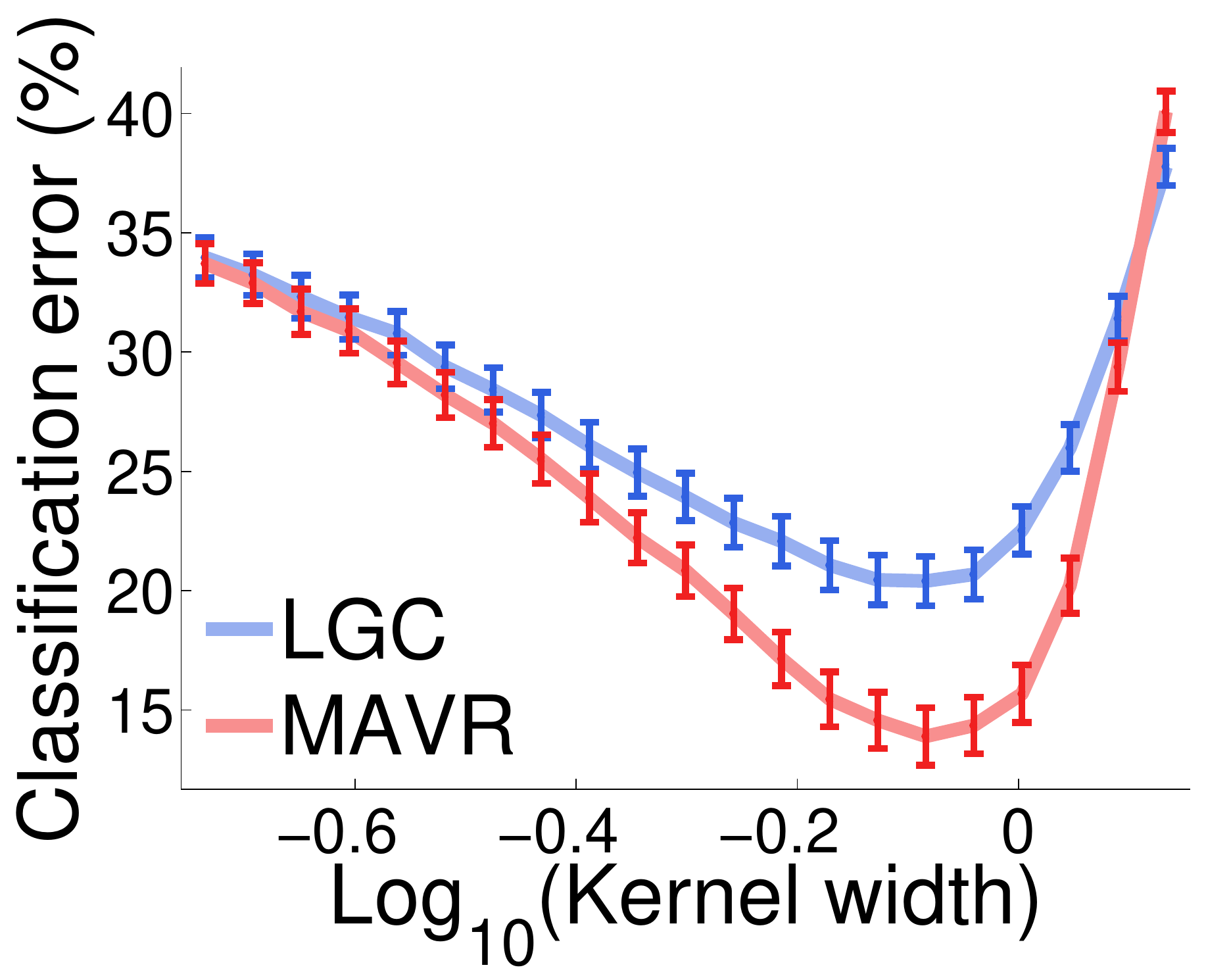}}~~~
  \subfigure[Varying $\gamma$]{\includegraphics[width=0.3\columnwidth]{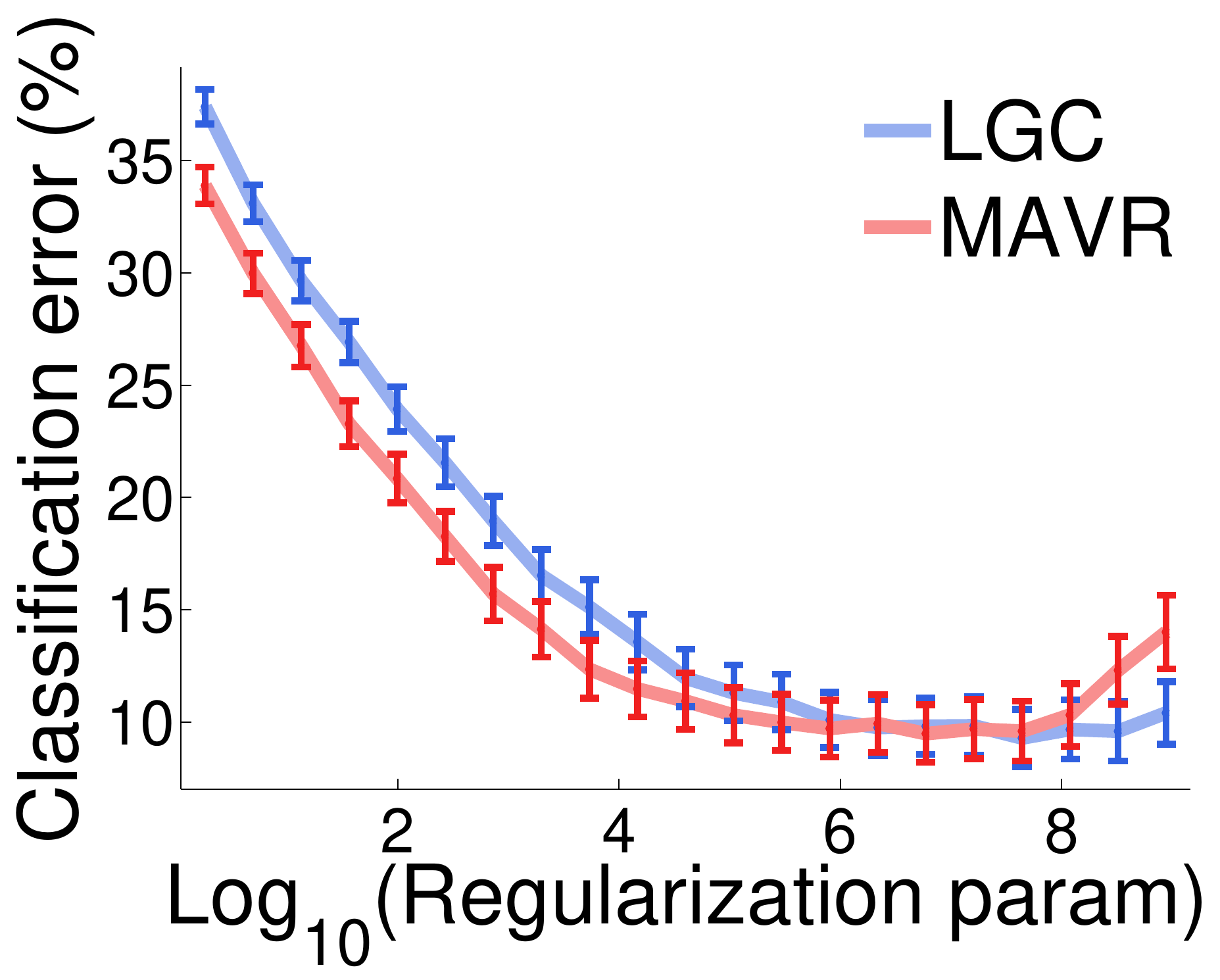}}\\
  \subfigure[Varying $\tau$]{\includegraphics[width=0.3\columnwidth]{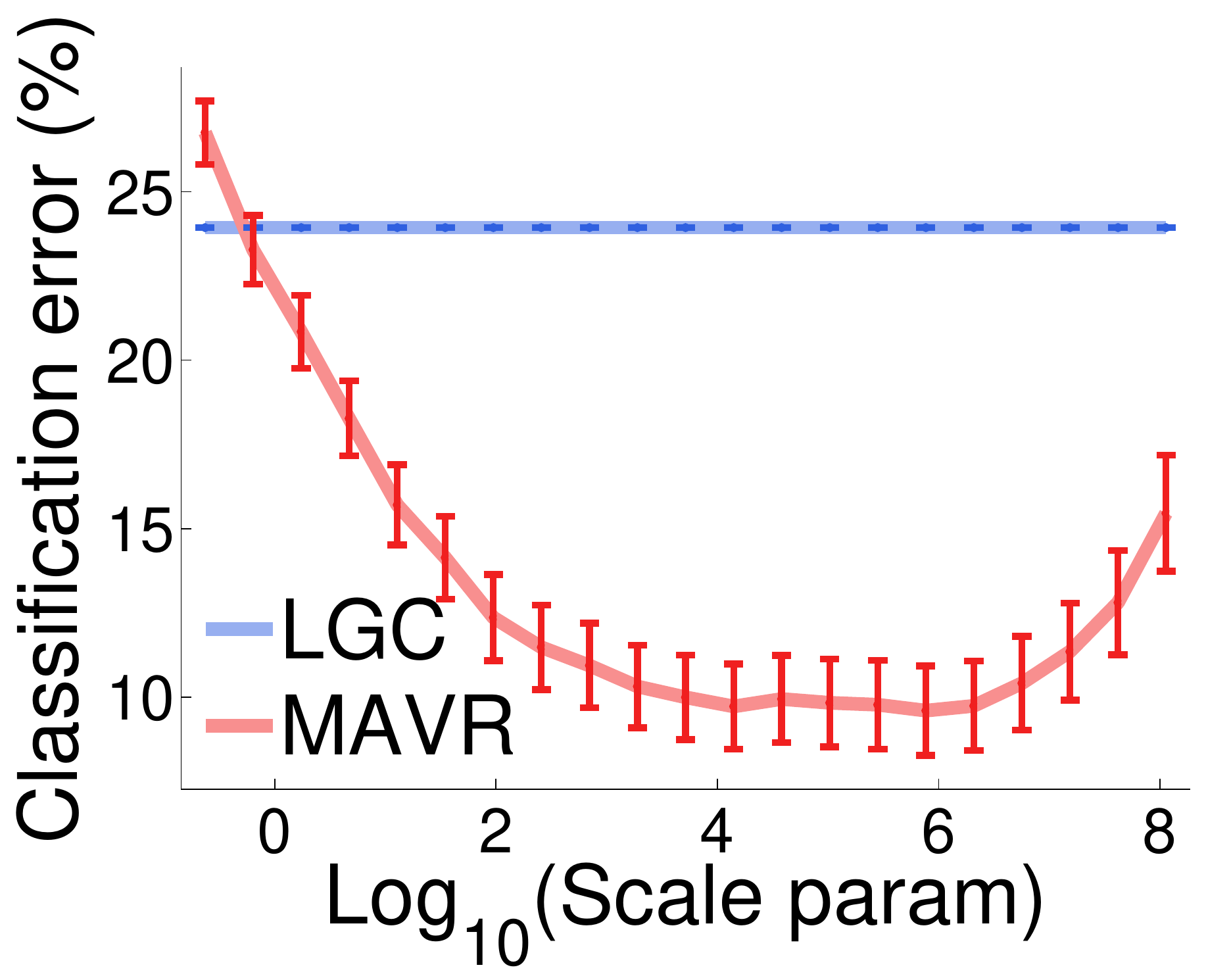}}~~~
  \subfigure[Varying $l$]{\includegraphics[width=0.3\columnwidth]{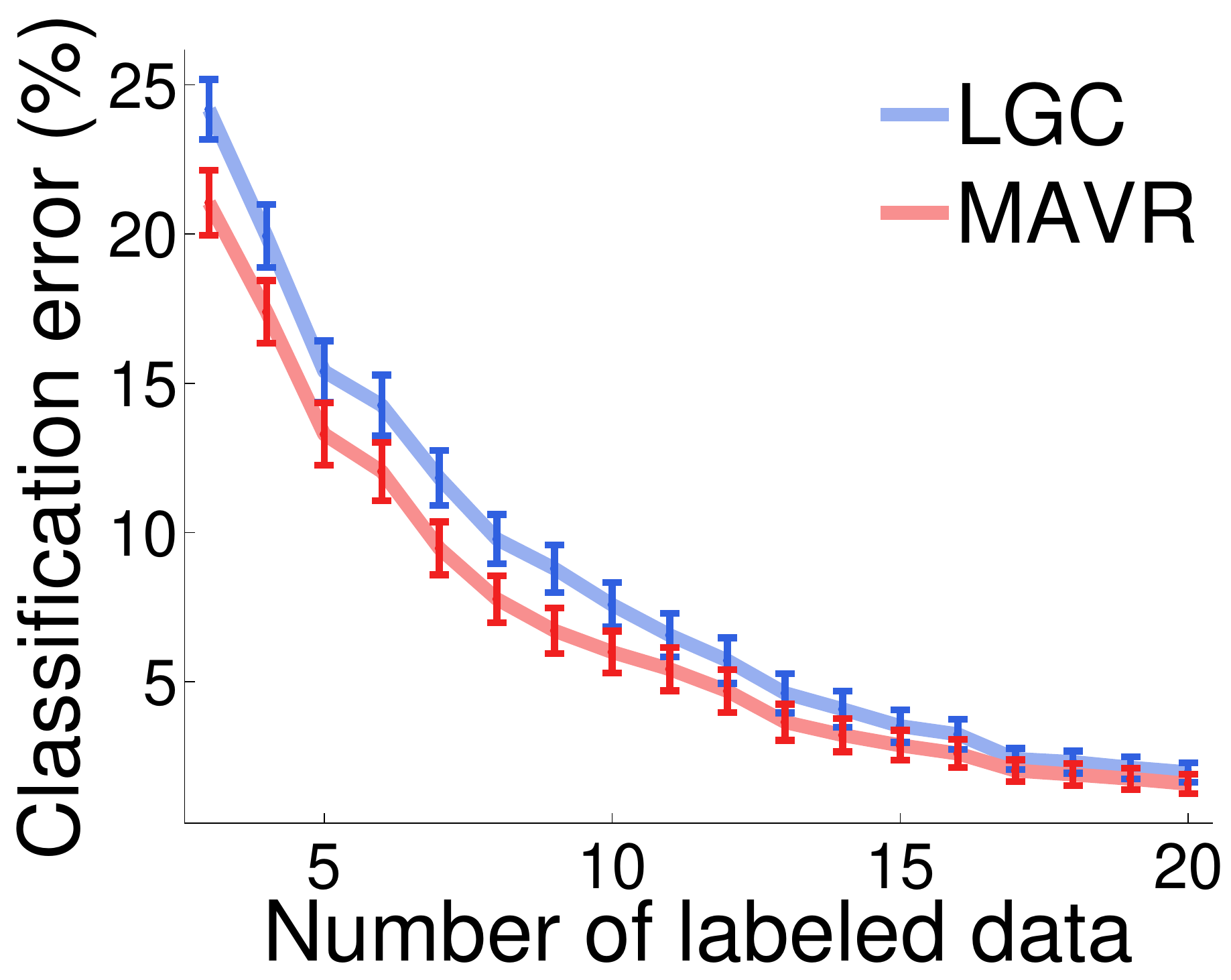}}~~~
  \subfigure[Varying $n$]{\includegraphics[width=0.3\columnwidth]{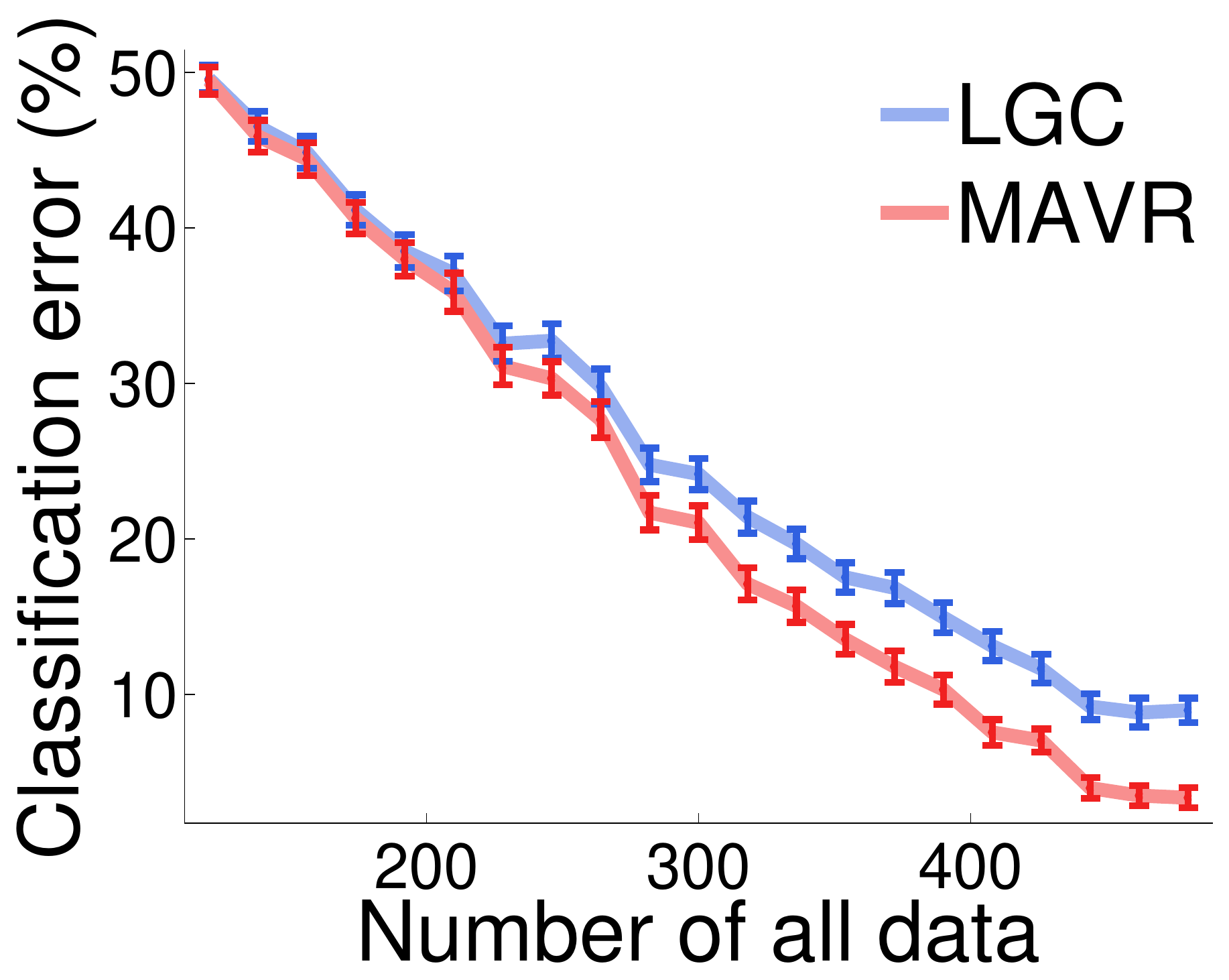}}
  \caption{Experimental results on the artificial data set 3circles. Means with standard errors are shown.}
  \label{fig:exp-3circles}
\end{figure*}

A state-of-the-art multi-class transductive learning method named \emph{learning with local and global consistency} (LGC) \citep{zhou03} is closely related to MAVR. Actually, if we specify $P=I_c$ and $Q=L_\textrm{nor}$, unconstrained MAVR will be reduced to LGC exactly. Although LGC is motivated by the label propagation viewpoint, it can be written as optimization (4) in \citet{zhou03}. Here, we illustrate the nuance of constrained MAVR and LGC that is unconstrained MAVR using an artificial data set.

The artificial data set \emph{3circles} is generated as follows. We have three classes with the class ratio $1/6,1/3,1/2$. Let $y_i$ be the ground-truth label of $x_i$, then $x_i$ is generated by
\[ x_i=(6y_i\cos(a_i)+\epsilon_{i,1},5y_i\sin(a_i)+\epsilon_{i,2})^\T\in\bR^2, \]
where $a_i$ is an angel drawn i.i.d.~from the uniform distribution $\cU(0,2\pi)$, and $\epsilon_{i,1}$ and $\epsilon_{i,2}$ are noises drawn i.i.d.~from the normal distribution $\cN(0,\sigma_\epsilon^2)$. We vary one factor and fix all other factors. The default values of these factors are $\sigma_\epsilon=0.5$, $\sigma=0.5$, $l=3$, $n=300$, $\gamma=99$, and $\tau=\sqrt{l}$. Figure~\ref{fig:exp-3circles} shows the experimental results, where the means with the standard errors of the classification error rates are plotted. For each task that corresponds to a full specification of all factors, MAVR and LGC were repeatedly ran on 100 random samplings. We can see from Figure~\ref{fig:exp-3circles} that the performance of LGC was usually not as good as MAVR.

\begin{figure*}[t]
  \centering
  \subfigure[USPS (Gaussian)]{\includegraphics[width=0.3\columnwidth]{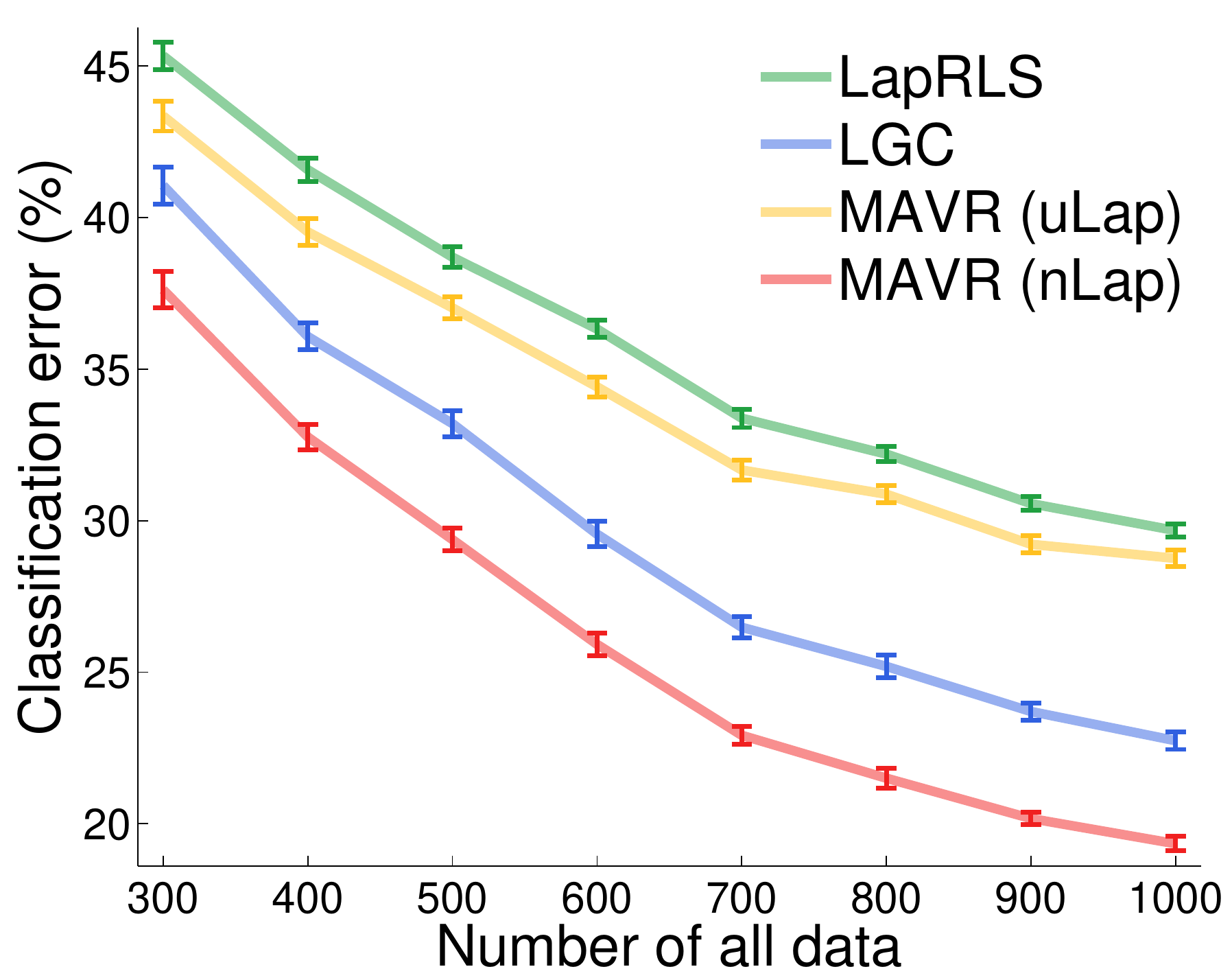}}~~~
  \subfigure[USPS (Local-scaling)]{\includegraphics[width=0.3\columnwidth]{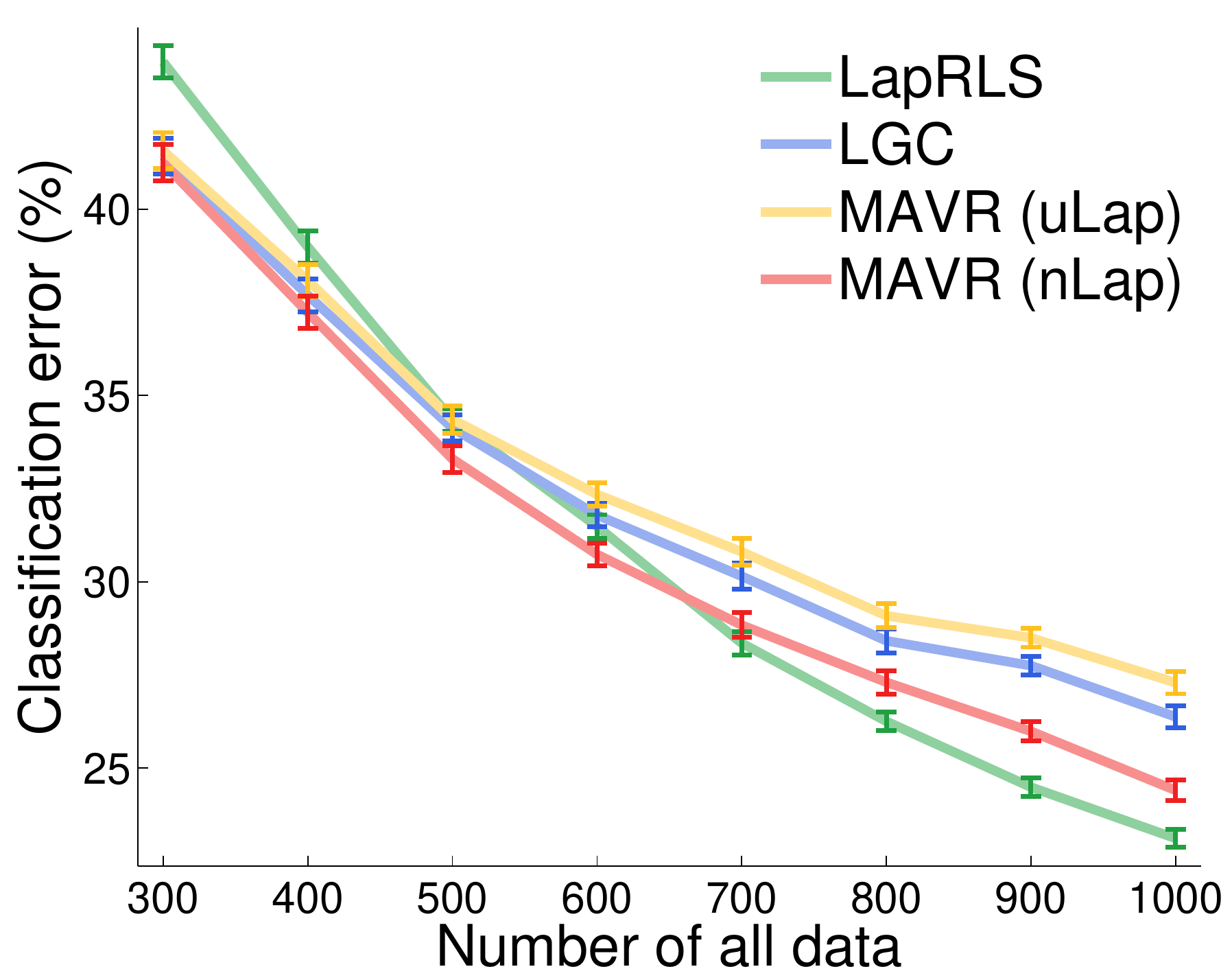}}~~~
  \subfigure[USPS (Cosine)]{\includegraphics[width=0.3\columnwidth]{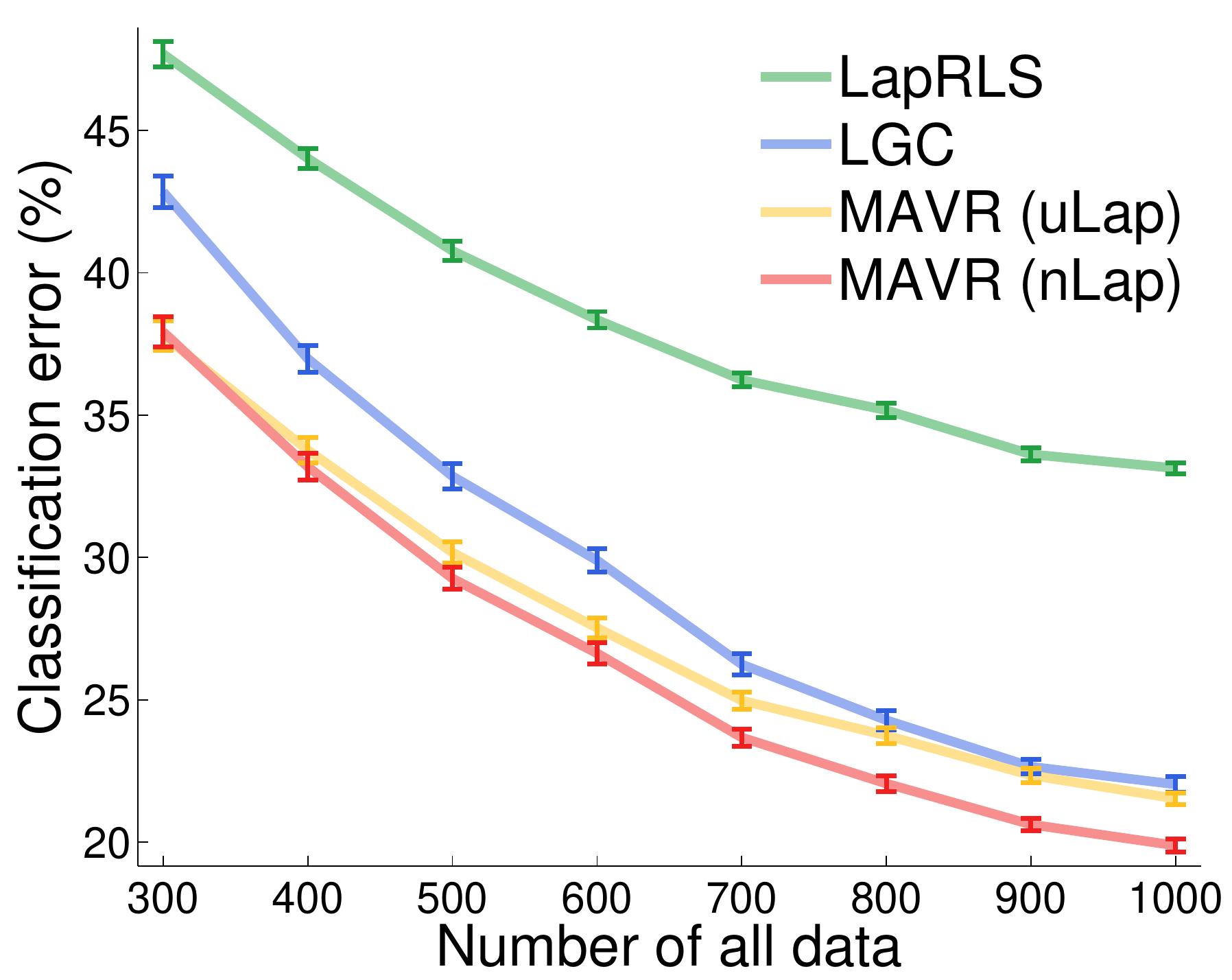}}\\
  \subfigure[MNIST (Gaussian)]{\includegraphics[width=0.3\columnwidth]{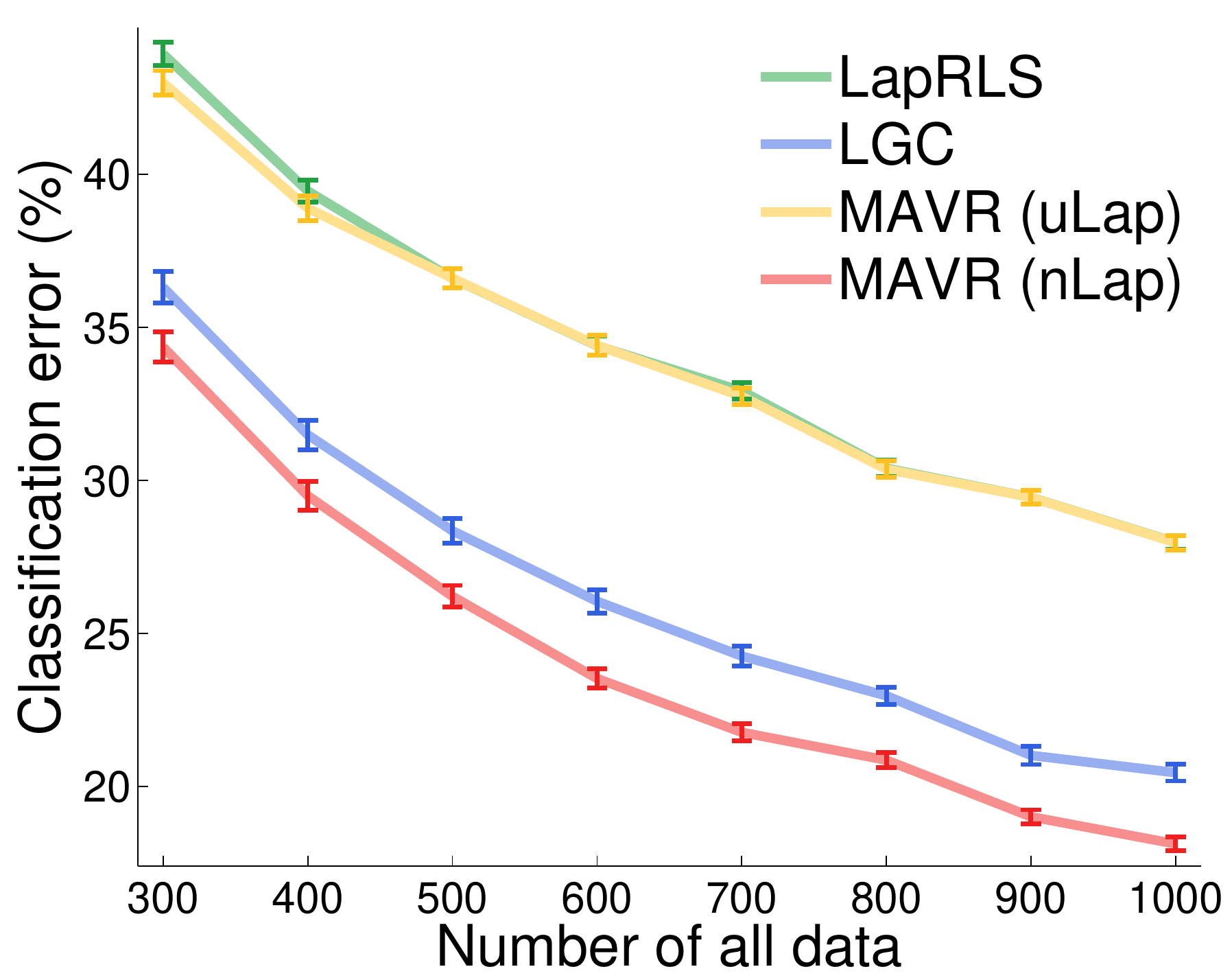}}~~~
  \subfigure[MNIST (Local-scaling)]{\includegraphics[width=0.3\columnwidth]{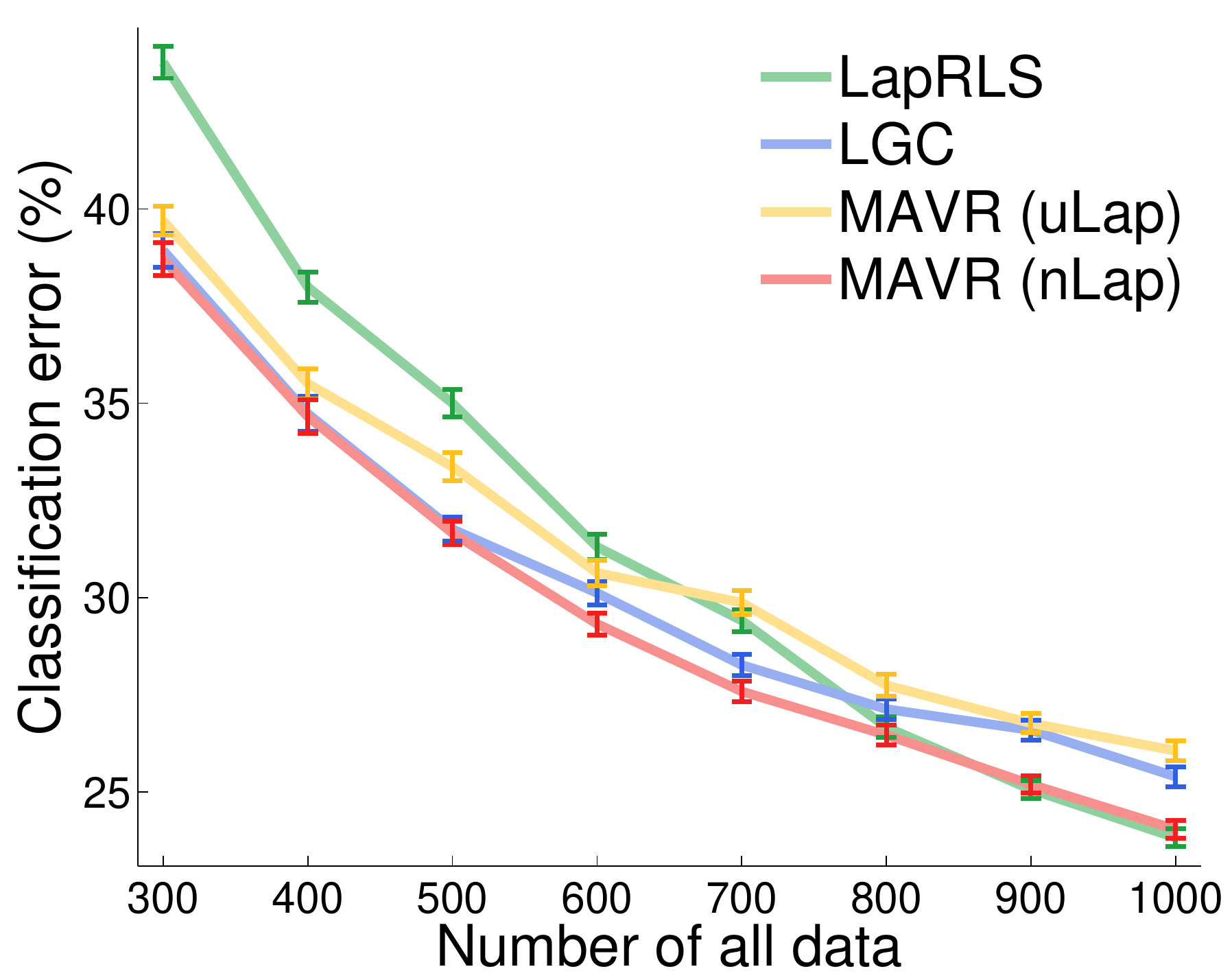}}~~~
  \subfigure[MNIST (Cosine)]{\includegraphics[width=0.3\columnwidth]{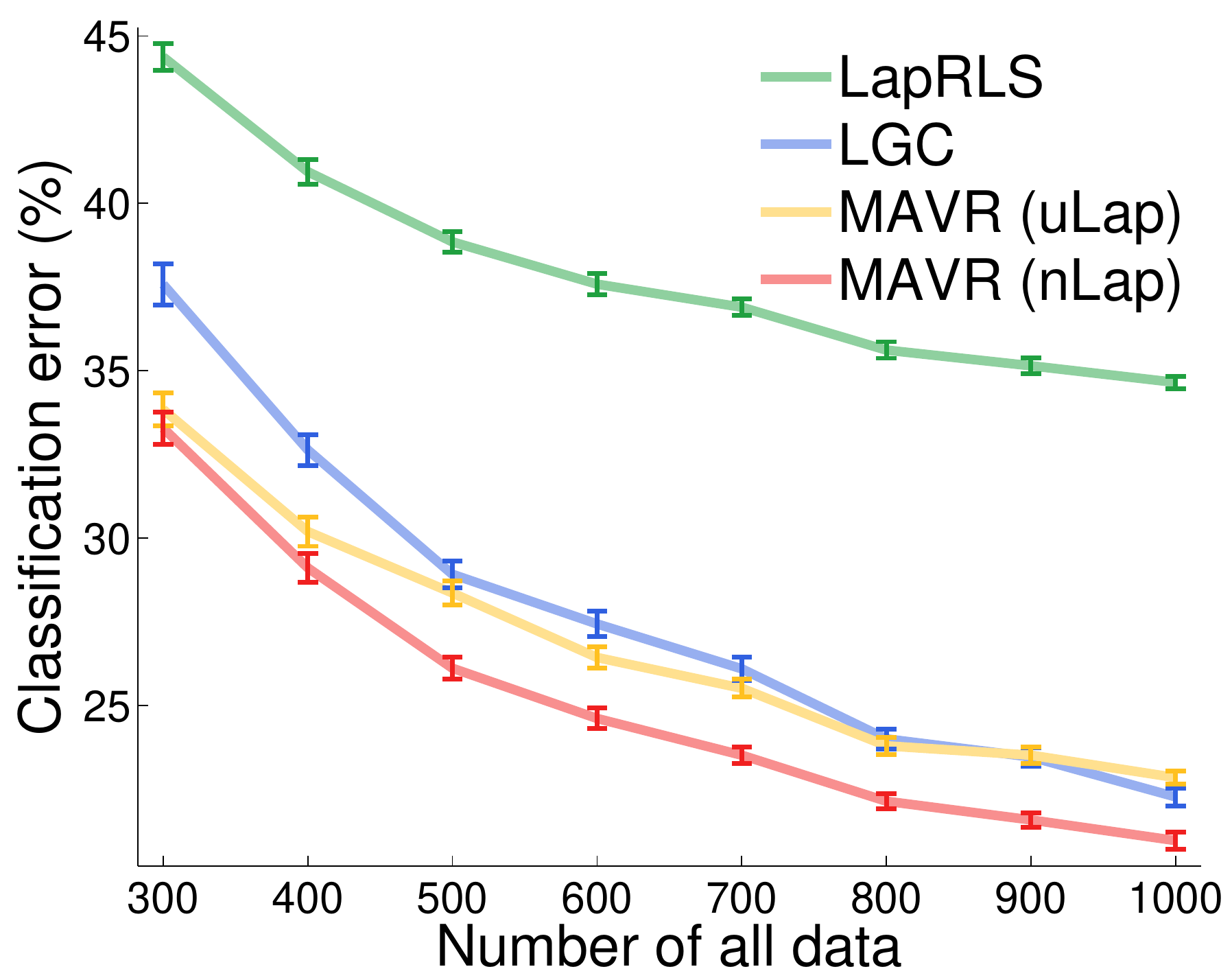}}\\
  \subfigure[20News (Gaussian)]{\includegraphics[width=0.3\columnwidth]{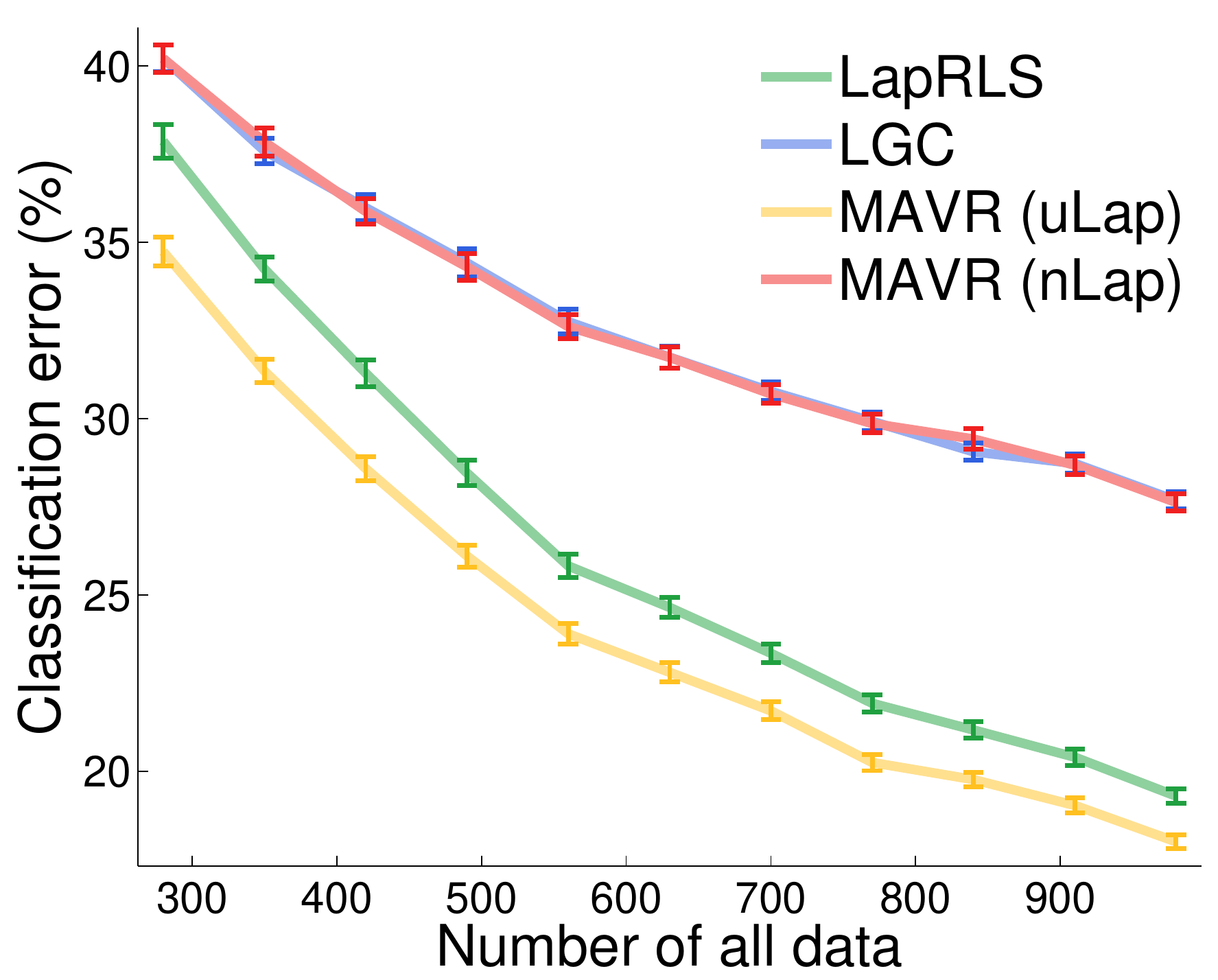}}~~~
  \subfigure[20News (Local-scaling)]{\includegraphics[width=0.3\columnwidth]{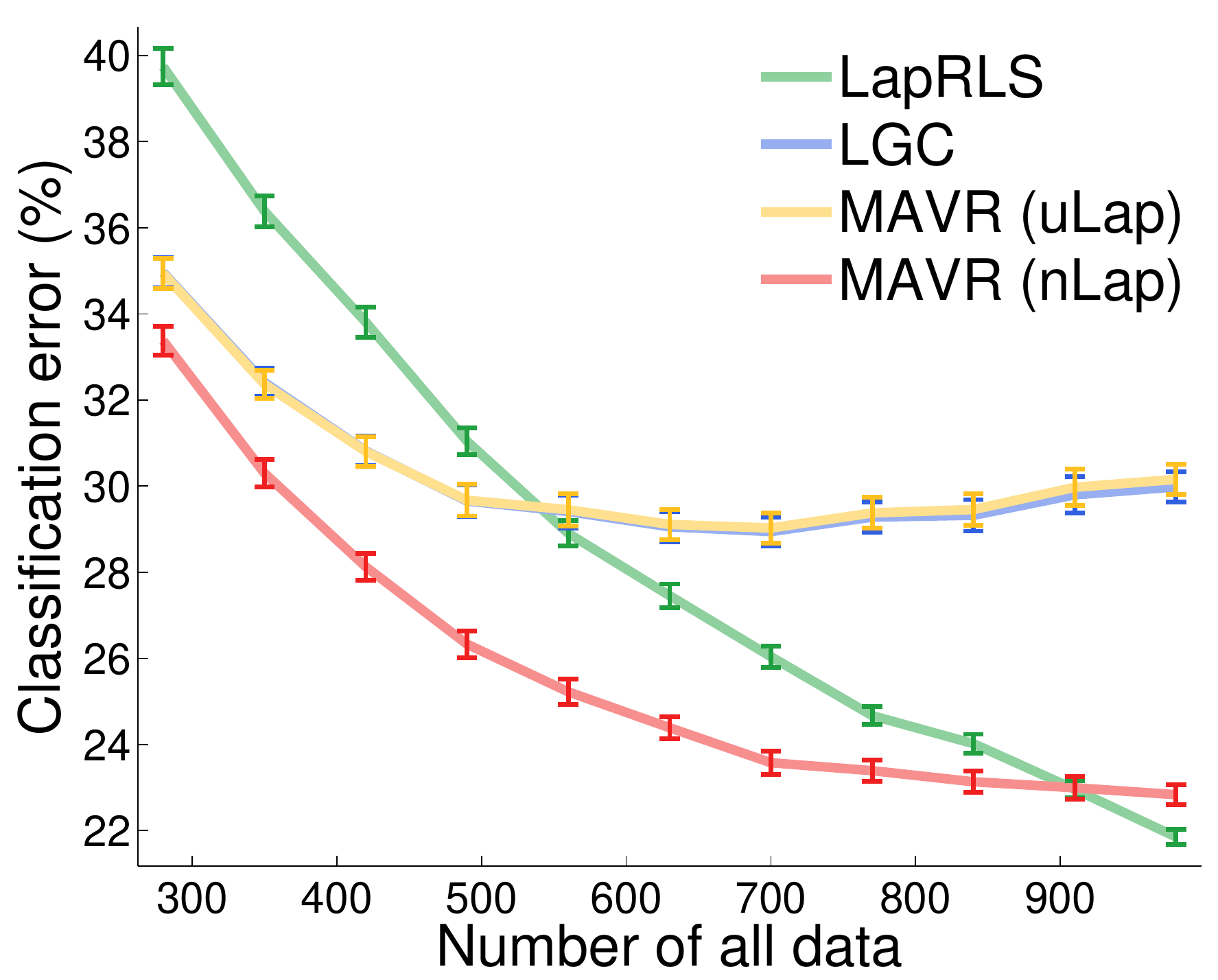}}~~~
  \subfigure[20News (Cosine)]{\includegraphics[width=0.3\columnwidth]{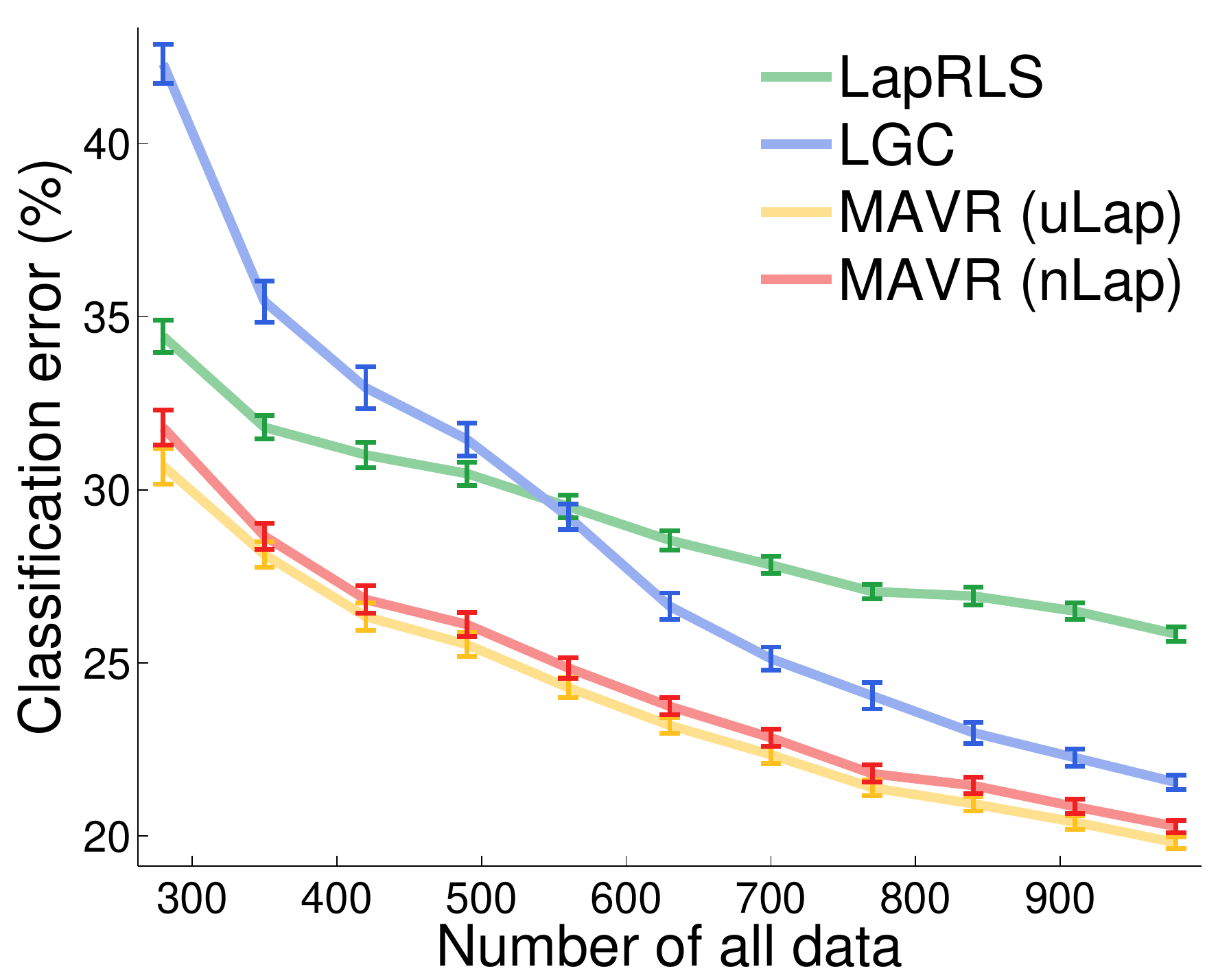}}\\
  \subfigure[Isolet (Gaussian)]{\includegraphics[width=0.3\columnwidth]{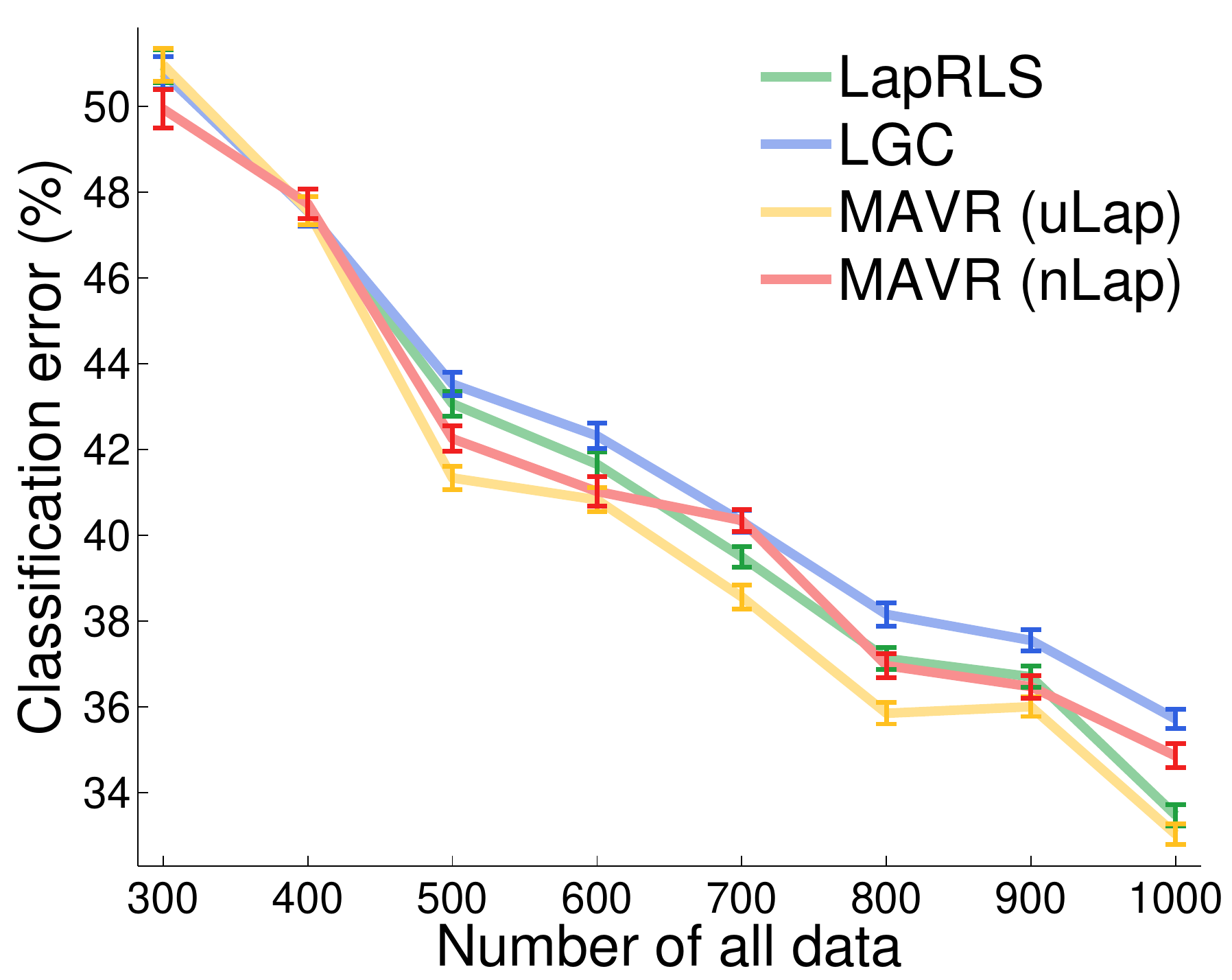}}~~~
  \subfigure[Isolet (Local-scaling)]{\includegraphics[width=0.3\columnwidth]{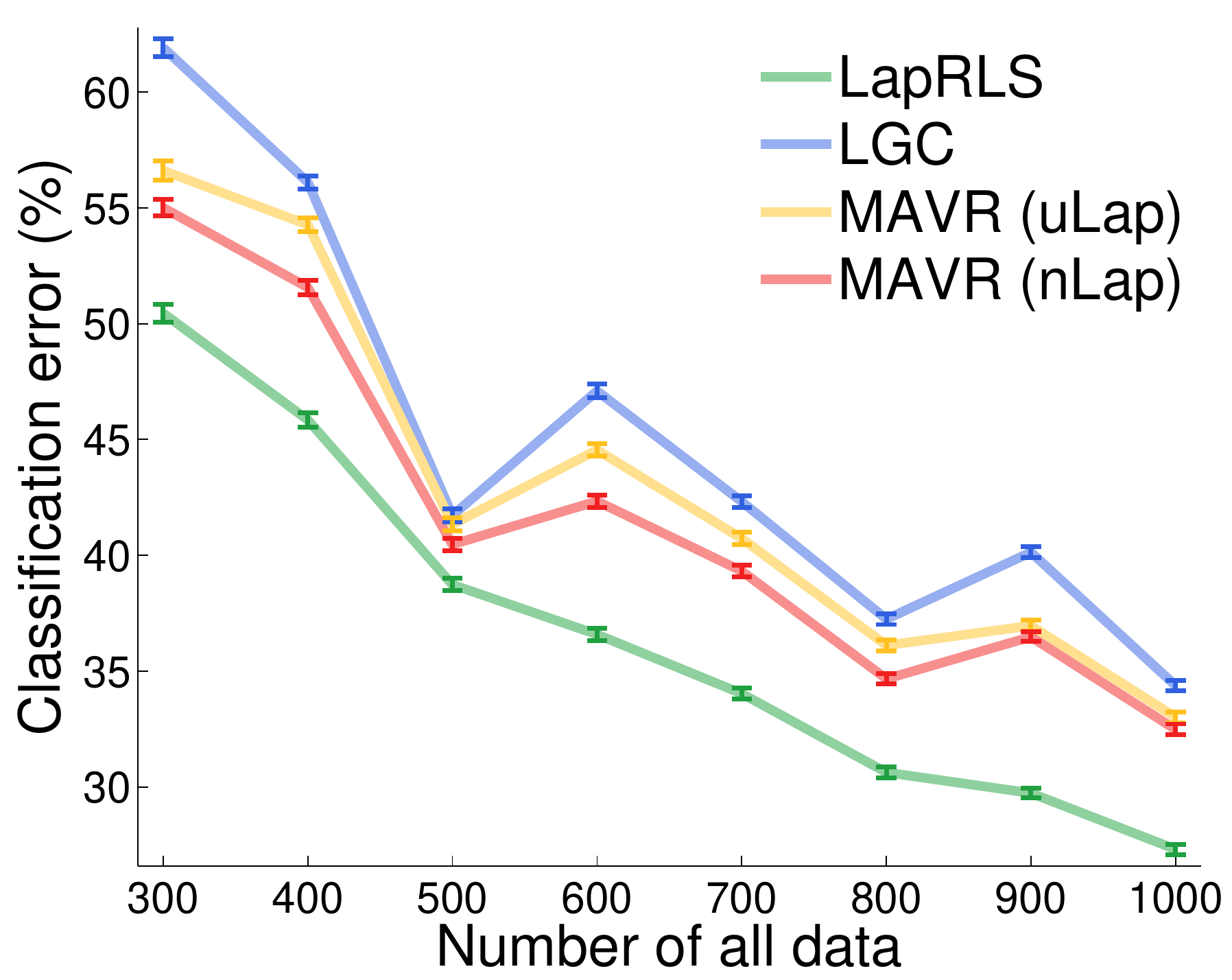}}~~~
  \subfigure[Isolet (Cosine)]{\includegraphics[width=0.3\columnwidth]{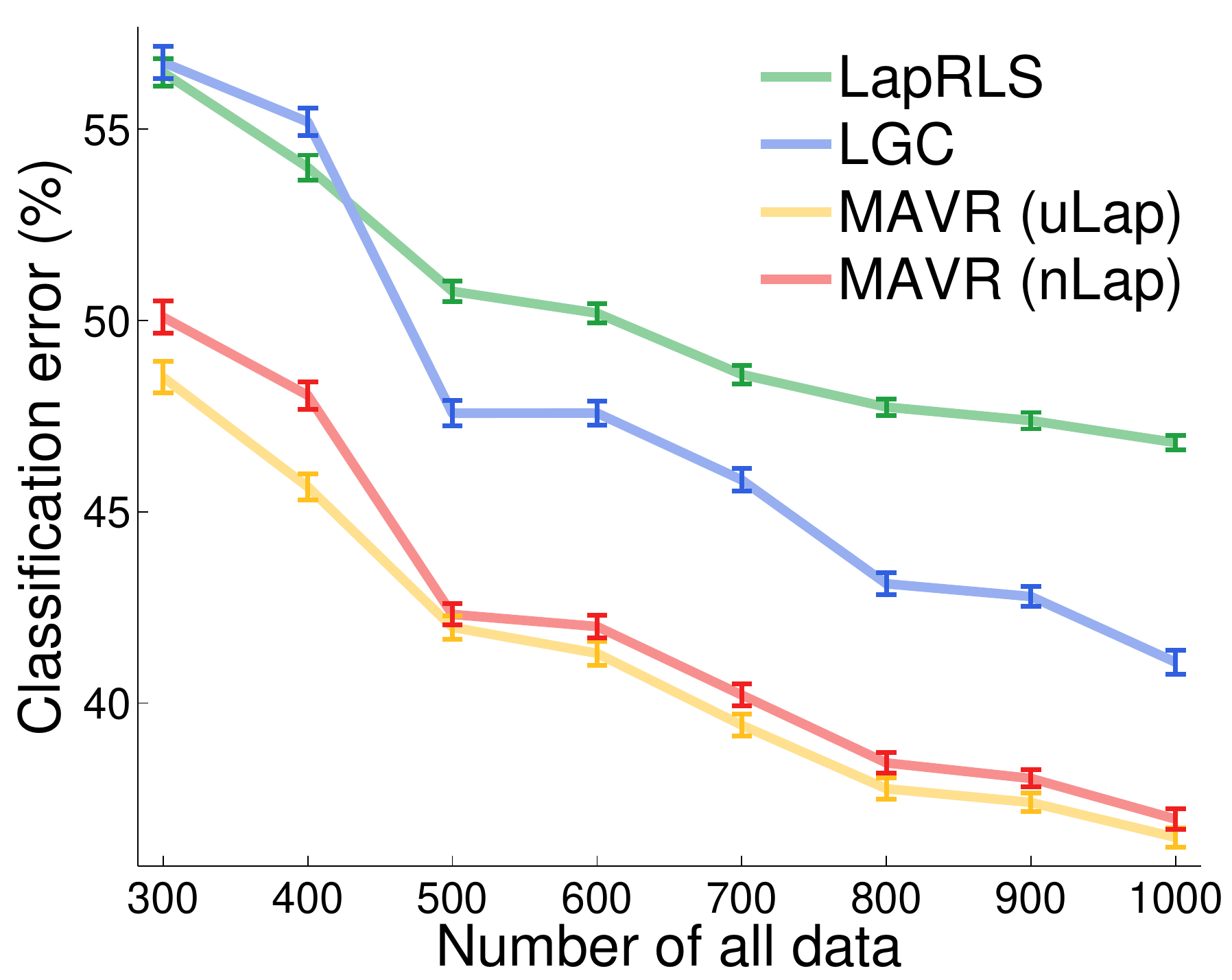}}
  \caption{Experimental results on USPS, MNIST, 20Newsgroups and Isolet. Means with standard errors are shown.}
  \label{fig:exp-benchmark}
\end{figure*}

Over the past decades, a huge number of transductive learning and semi-supervised learning methods have been proposed based on various motivations as graph cut \citep{blum01}, random walk \citep{zhu03}, manifold regularization \citep{belkin06}, and information maximization \citep{niu13b}, just to name a few. A state-of-the-art semi-supervised learning method called \emph{Laplacian regularized least squares} (LapRLS) \citep{belkin06} is included to be compared with MAVR besides LGC.

The experimental results are reported in Figure~\ref{fig:exp-benchmark}. Similarly to Figure~\ref{fig:exp-3circles}, the means with the standard errors of the classification error rates are shown where 4 methods were repeatedly ran on 100 random samplings for each task. We considered another specification of $Q$ as the \emph{unnormalized graph Laplacian} $L_\textrm{un}=D-W$ which was also employed by LapRLS. The cosine similarity is defined by
\[ W_{i,j}=x_i^\T x_j/\|x_i\|_2\|x_j\|_2\textrm{ if }x_i\sim_kx_j,\quad W_{i,j}=0\textrm{ otherwise}, \]
where $x_i\sim_kx_j$ means $x_i$ and $x_j$ are among the $k$-nearest neighbors of each other. We set $l=n/10$ for all involved $n$ in Figure~\ref{fig:exp-benchmark}, and there seems no reliable model selection method given very few labeled data, so we select the best hyperparameters for each method using the labels of unlabeled data from 10 additional random samplings. Specifically, $\sigma$ is the median distance $\times$ $\{1/16,1/8,1/4,1/2,1\}$, and $k$ is from $\{1,3,5,7,9\}$ for both local-scaling and cosine similarities; $\tau$ is $\sqrt{l}\times\{1/16,1/8,1/4,1/2,1\}$. The hyperparameters are all fixed since it resulted in more stable performance. For MAVR, LGC, and $\lambda_I$ of LapRLS, it was fixed to 99 if the Gaussian and cosine similarities were used and 1 if the local-scaling similarity was used; $\lambda_A$ of LapRLS was $10^{-3}$ if the Gaussian and local-scaling similarities were used and $10^3$ if the cosine similarity was used since LapRLS also needed $W$ that was too sparse and near singular, but an exception was panel (i) where $\lambda_A=10^{-3}$ gave lower error rates of LapRLS. We can see from Figure~\ref{fig:exp-benchmark} that two MAVR methods often compared favorably with the state-of-the-art methods LGC and LapRLS, which implies that our proposed multi-class volume approximation is reasonable and practical.

\section{Conclusions}

We proposed a multi-class volume approximation that can be applied to several transductive problem settings such as multi-class, multi-label and serendipitous learning. The resultant learning method is non-convex in nature but can be solved exactly and efficiently. It is theoretically justified by our stability and error analyses and experimentally demonstrated promising.

\appendix
\section{Proofs}
\label{app:proofs}%

\subsection{Proof of Theorem~\ref{thm:opt-rho}}

The derivative of $g(\rho)$ is
\[ g'(\rho)=\sum_{k=k_0}^{nc}\frac{2z_k^2}{(\gamma\lambda_{PQ,k}-\rho)^3}-\tau^2. \]
Hence, $g'(\rho)>0$ whenever $\rho<\gamma\lambda_{PQ,k_0}$, and $g(\rho)$ is strictly increasing in the interval $(-\infty,\gamma\lambda_{PQ,k_0})$. Moreover,
\[ \lim_{\rho\to-\infty}g(\rho)=-\tau^2 \quad\textrm{and}\quad \lim_{\rho\to\gamma\lambda_{PQ,k_0}}g(\rho)=+\infty, \]
and thus $g(\rho)$ has exactly one root in $(-\infty,\gamma\lambda_{PQ,k_0})$. Notice that $\|\vz\|_2=\|\vect(V_Q^\T YV_P)\|_2=\|V_{PQ}^\T\vy\|_2=\|\vy\|_2$ since $V_{PQ}$ is an orthonormal matrix, and then $\rho_0=\gamma\lambda_{PQ,k_0}-\|\vy\|_2/\tau=\gamma\lambda_{PQ,k_0}-\|\vz\|_2/\tau$. As a result,
\begin{align*}
g(\rho_0) &= \sum_{k=k_0}^{nc}\frac{z_k^2}{(\gamma\lambda_{PQ,k}-\rho_0)^2}-\tau^2\\
&= \sum_{k=k_0}^{nc}\frac{z_k^2}{(\gamma\lambda_{PQ,k}-\gamma\lambda_{PQ,k_0}+\|\vz\|_2/\tau)^2}-\tau^2\\
&\le \sum_{k=k_0}^{nc}\frac{z_k^2}{(\|\vz\|_2/\tau)^2}-\tau^2\\
&= \left(\frac{\sum_{k=k_0}^{nc}z_k^2}{\|\vz\|_2^2}-1\right)\tau^2\\
&\le 0,
\end{align*}
where the first inequality is because $\lambda_{PQ,k}\ge\lambda_{PQ,k_0}$ for $k\ge k_0$. The fact that $g(\rho_0)\le0$ concludes that the only root in $(-\infty,\gamma\lambda_{PQ,k_0})$ is in $[\rho_0,\gamma\lambda_{PQ,k_0})$ but not $(-\infty,\rho_0)$. \qed

\subsection{Proof of Theorem~\ref{thm:stable}}

Denote by $\vh=\vect(H)$, $\vy=\vect(Y)$ and $M=(\gamma P\otimes Q-\rho I_{nc})$, and denote by $\vh'$, $\vy'$ and $M'$ similarly. Let $\lambda_{\min}(\cdot)$ and $\lambda_{\max}(\cdot)$ be two functions extracting the smallest and largest eigenvalues of a matrix. Under our assumption,
\[ \lambda_{\min}(M)=\gamma\lambda_{PQ,1}-\rho\ge C_{\gamma,\tau}>0 \]
which means that $M$ is positive definite, and so is $M'$. By Eq.~\eqref{eq:opt-h},
\begin{align*}
\vh-\vh' &= M^{-1}\vy-M'^{-1}\vy'\\
&= M^{-1}(\vy-\vy')+(M^{-1}-M'^{-1})\vy'\\
&= M^{-1}(\vy-\vy')+M^{-1}(M'-M)M'^{-1}\vy'\\
&= M^{-1}(\vy-\vy')+(\rho'-\rho)M^{-1}M'^{-1}\vy'.
\end{align*}
Note that $\|A\vv\|_2\le\lambda_{\max}(A)\|\vv\|_2$ for any symmetric positive-definite matrix $A$ and any vector $\vv$, as well as $\lambda_{\max}(AB)\le\lambda_{\max}(A)\lambda_{\max}(B)$ for any symmetric positive-definite matrices $A$ and $B$. Hence,
\begin{align*}
\|\vh-\vh'\|_2 &= \|M^{-1}(\vy-\vy')+(\rho'-\rho)M^{-1}M'^{-1}\vy'\|_2\\
&\le \|M^{-1}(\vy-\vy')\|_2+|\rho-\rho'|\|M^{-1}M'^{-1}\vy'\|_2\\
&\le \lambda_{\max}(M^{-1})\|\vy-\vy'\|_2+\lambda_{\max}(M^{-1})\lambda_{\max}(M'^{-1})|\rho-\rho'|\|\vy'\|_2\\
&\le \frac{\|\vy-\vy'\|_2}{C_{\gamma,\tau}}+\frac{|\rho-\rho'|\|\vy'\|_2}{C_{\gamma,\tau}^2},
\end{align*}
where the first inequality is the triangle inequality, the second inequality is because $M^{-1}$ and $M'^{-1}$ are symmetric positive definite, and the third inequality follows from $\lambda_{\max}(M^{-1})=1/\lambda_{\min}(M)$ and $\lambda_{\max}(M'^{-1})=1/\lambda_{\min}(M')$. Due to the symmetry of $\vh$ and $\vh'$,
\[ \|\vh-\vh'\|_2 \le \frac{\|\vy-\vy'\|_2}{C_{\gamma,\tau}}
+\frac{|\rho-\rho'|\min\{\|\vy\|_2,\|\vy'\|_2\}}{C_{\gamma,\tau}^2}. \]
This inequality is the vectorization of \eqref{eq:stable}.

For MAVR in optimization \eqref{eq:mavr}, Theorem~\ref{thm:opt-rho} together with our assumption indicates that
\begin{align*}
\gamma\lambda_{PQ,1}-\|\vy\|_2/\tau &\le \rho<\gamma\lambda_{PQ,1},\\
\gamma\lambda_{PQ,1}-\|\vy'\|_2/\tau &\le \rho'<\gamma\lambda_{PQ,1},
\end{align*}
so $|\rho'-\rho|\le\max\{\|\vy\|_2/\tau,\|\vy'\|_2/\tau\}$ and
\begin{align*}
\|\vh-\vh'\|_2 &\le \frac{\|\vy-\vy'\|_2}{C_{\gamma,\tau}}
+\frac{\max\{\|\vy\|_2,\|\vy'\|_2\}\min\{\|\vy\|_2,\|\vy'\|_2\}}{\tau C_{\gamma,\tau}^2}\\
&= \frac{\|\vy-\vy'\|_2}{C_{\gamma,\tau}}
+\frac{\|\vy\|_2\|\vy'\|_2}{\tau C_{\gamma,\tau}^2}.
\end{align*}
For unconstrained MAVR in optimization \eqref{eq:mavr-uncon}, we have
\[ \|\vh-\vh'\|_2 \le \frac{\|\vy-\vy'\|_2}{C_{\gamma,\tau}}, \]
since $\rho=\rho'=-1$. \qed

\subsection{Proof of Theorem~\ref{thm:error}}

Denote by $\vh=\vect(H)$, $\vy=\vect(Y)$, $\vh^*=\vect(H^*)$, $\ve=\vect(E)$, and $M=\gamma P\otimes Q$. The Kronecker product $P\otimes Q$ is symmetric and positive definite, and then $M^{1/2}$ is a well-defined symmetric and positive-definite matrix. We can know based on $V(H^*)\le C_h$ that
\[ \|M^{1/2}\vh^*\|_2 = \sqrt{\gamma\vh^{*^\T}(P\otimes Q)\vh^*}
\le \sqrt{\gamma C_h\|\vh^*\|_2^2} = \sqrt{\gamma C_h}\|\vh^*\|_2. \]
Let $\lambda_{\min}(\cdot)$ and $\lambda_{\max}(\cdot)$ be two functions extracting the smallest and largest eigenvalues of a matrix. In the following, we will frequently use that $\|A\vv\|_2\le\lambda_{\max}(A)\|\vv\|_2$ for any symmetric positive-definite matrix $A$ and any vector $\vv$.

Consider unconstrained MAVR in optimization \eqref{eq:mavr-uncon} first. Since $\rho=-1$,
\begin{align*}
\vh-\vh^* &= (M+I_{nc})^{-1}\vy-\vh^*\\
&= (M+I_{nc})^{-1}(\vh^*+\ve)-(M+I_{nc})^{-1}(M+I_{nc})\vh^*\\
&= -(M+I_{nc})^{-1}M\vh^*+(M+I_{nc})^{-1}\ve.
\end{align*}
As a consequence,
\[ \bE\|\vh-\vh^*\|_2^2=\|(M+I_{nc})^{-1}M\vh^*\|_2^2+\bE\|(M+I_{nc})^{-1}\ve\|_2^2, \]
since $\bE[(M+I_{nc})^{-1}\ve]=(M+I_{nc})^{-1}\bE\ve=\zero_{nc}$. Subsequently,
\begin{align*}
\|(M+I_{nc})^{-1}M\vh^*\|_2 &\le \lambda_{\max}((M+I_{nc})^{-1}M^{1/2})\cdot\|M^{1/2}\vh^*\|_2\\
&\le \lambda_{\max}((\gamma P\otimes Q+I_{nc})^{-1}(\gamma P\otimes Q)^{1/2})\cdot\sqrt{\gamma C_h}\|\vh^*\|_2\\
&= \sqrt{\gamma C_h}\lambda_{\max}\left(\frac{\sqrt{\gamma}}{\gamma+1}
(\Lambda_{PQ}+I_{nc})^{-1}\Lambda_{PQ}^{1/2}\right)\|\vh^*\|_2\\
&\le \sqrt{C_h}\lambda_{\max}((\Lambda_{PQ}+I_{nc})^{-1}\Lambda_{PQ}^{1/2})\|\vh^*\|_2\\
&\le \frac{1}{2}\sqrt{C_h}\|\vh^*\|_2,
\end{align*}
where the last inequality is because the eigenvalues of $(\Lambda_{PQ}+I_{nc})^{-1}\Lambda_{PQ}^{1/2}$ are $\frac{\sqrt{\lambda_{PQ,1}}}{\lambda_{PQ,1}+1},\ldots,\frac{\sqrt{\lambda_{PQ,nc}}}{\lambda_{PQ,nc}+1}$ and
\[ \sup\nolimits_{\lambda\ge0} \frac{\sqrt{\lambda}}{\lambda+1}=\frac{1}{2}. \]
On the other hand,
\begin{align*}
\bE\|(M+I_{nc})^{-1}\ve\|_2^2 &\le (\lambda_{\max}((M+I_{nc})^{-1}))^2\cdot\bE\|\ve\|_2^2\\
&= \frac{\bE[\ve^\T\ve]}{(\lambda_{\min}(M+I_{nc}))^2}\\
&\le \widetilde{l}\sigma_l^2+\widetilde{u}\sigma_u^2.
\end{align*}
Hence,
\[ \bE\|\vh-\vh^*\|_2^2 \le \frac{1}{4}C_h\|\vh^*\|_2^2+\widetilde{l}\sigma_l^2+\widetilde{u}\sigma_u^2, \]
which completes the proof of inequality \eqref{eq:error-uncon}.

Next, consider MAVR in optimization \eqref{eq:mavr}. We would have
\begin{align*}
\vh-\vh^* &= (M-\rho I_{nc})^{-1}\vy-\vh^*\\
&= (M-\rho I_{nc})^{-1}(\vh^*+\ve)-(M-\rho I_{nc})^{-1}(M-\rho I_{nc})\vh^*\\
&= -(M-\rho I_{nc})^{-1}(M-(\rho+1)I_{nc})\vh^*+(M-\rho I_{nc})^{-1}\ve.
\end{align*}
In general, $\bE[(M-\rho I_{nc})^{-1}\ve]\neq\zero_{nc}$ since $\rho$ depends on $\ve$. Furthermore, $M-(\rho+1)I_{nc}$ may have negative eigenvalues when $\gamma\lambda_{PQ,1}-1<\rho\le\gamma\lambda_{PQ,1}-C_{\gamma,\tau}$. Taking the expectation of $\|\vh-\vh^*\|_2$,
\begin{align*}
\bE\|\vh-\vh^*\|_2 &\le \bE\|(M-\rho I_{nc})^{-1}(M-(\rho+1)I_{nc})\vh^*\|_2+\bE\|(M-\rho I_{nc})^{-1}\ve\|_2\\
&\le \bE\|(M-\rho I_{nc})^{-1}M\vh^*\|_2+\bE[|\rho+1|\|(M-\rho I_{nc})^{-1}\vh^*\|_2]+\bE\|(M-\rho I_{nc})^{-1}\ve\|_2.
\end{align*}
Subsequently,
\begin{align*}
\bE\|(M-\rho I_{nc})^{-1}M\vh^*\|_2
&\le \sup\nolimits_\rho \lambda_{\max}((M-\rho I_{nc})^{-1}M^{1/2})\cdot\sqrt{\gamma C_h}\|\vh^*\|_2\\
&= \sup\nolimits_\rho \sqrt{C_h}\lambda_{\max}\left((\Lambda_{PQ}-\rho/\gamma I_{nc})^{-1}
\Lambda_{PQ}^{1/2}\right)\|\vh^*\|_2\\
&\le \sqrt{C_h}\|\vh^*\|_2\cdot\sup\nolimits_{\rho\le\gamma\lambda_{PQ,1}-C_{\gamma,\tau}}
\sup\nolimits_{\lambda\ge\lambda_{PQ,1}}\left(\frac{\sqrt{\lambda}}{\lambda-\rho/\gamma}\right)\\
&\le \frac{\sqrt{C_h}\gamma\lambda_{PQ,1}}{C_{\gamma,\tau}}\|\vh^*\|_2.
\end{align*}
On the other hand,
\begin{align*}
\bE[|\rho+1|\|(M-\rho I_{nc})^{-1}\vh^*\|_2]
&\le \bE|\rho+1|\cdot\sup\nolimits_\rho\lambda_{\max}((M-\rho I_{nc})^{-1})\|\vh^*\|_2\\
&\le \frac{\|\vh^*\|_2}{C_{\gamma,\tau}}\cdot
\bE\max\{-\rho-1,\sup\nolimits_\rho\rho+1\}\\
&\le \frac{\|\vh^*\|_2}{C_{\gamma,\tau}}\cdot
\max\{\bE\|\vy\|_2/\tau-\gamma\lambda_{PQ,1}-1,\gamma\lambda_{PQ,1}-C_{\gamma,\tau}+1\}\\
&= \frac{\|\vh^*\|_2}{C_{\gamma,\tau}}\cdot\max\{\sqrt{\widetilde{l}}/\tau
-\gamma\lambda_{PQ,1}-1,\gamma\lambda_{PQ,1}-C_{\gamma,\tau}+1\}.
\end{align*}
where we used the fact that $\sup\nolimits_\rho\rho$ is independent of $\ve$, and applied \emph{Jensen's inequality} to obtain that
\[ \bE\|\vy\|_2\le\sqrt{\bE\|\vy\|_2^2}\le\sqrt{\widetilde{l}}. \]
In the end,
\begin{align*}
\bE\|(M-\rho I_{nc})^{-1}\ve\|_2
&\le \sup\nolimits_\rho\lambda_{\max}((M-\rho I_{nc})^{-1})\cdot\bE\|\ve\|_2\\
&\le \frac{\bE\sqrt{\ve^\T\ve}}{C_{\gamma,\tau}}\\
&\le \frac{\sqrt{\bE[\ve^\T\ve]}}{C_{\gamma,\tau}}\\
&= \frac{\sqrt{\widetilde{l}\sigma_l^2+\widetilde{u}\sigma_u^2}}{C_{\gamma,\tau}},
\end{align*}
where the third inequality is due to Jensen's inequality. Therefore, inequality \eqref{eq:error-con} follows by combining the three upper bounds of expectations. \qed

\bibliography{mavr_short}

\end{document}